\documentclass[10pt,twocolumn,letterpaper]{article}

\usepackage{cvpr}
\usepackage{times}
\usepackage{epsfig}
\usepackage{graphicx}
\usepackage{amsmath}
\usepackage{amssymb}
\usepackage{xcolor}
\usepackage{booktabs}
\usepackage{multirow}
\usepackage{subcaption}
\usepackage{bbm}

\usepackage{empheq}  

\newcommand{\trace}{\operatorname{tr}}
\renewcommand{\Re}{\operatorname{Re}}

\newcommand{\R}{\mathbb{R}}

\newcommand{\ones}{\mathbbm{1}}

\newcommand{\cfnet}{CFNet\xspace}

\usepackage[pagebackref=true,breaklinks=true,letterpaper=true,colorlinks,bookmarks=false]{hyperref}

\cvprfinalcopy 

\def\cvprPaperID{1027} 

\ifcvprfinal\fi
\begin{document}
\newcommand{\jack}[1]{{\color{blue} [Jack: #1]}}
\newcommand{\luca}[1]{{\color{orange} [Luca: #1]}}
\newcommand{\joao}[1]{{\color{green!50!black} [Joao: #1]}}
\newcommand{\todo}[1]{{\color{red} [TODO: #1]}}
\newcommand{\tocite}[1]{{\color{purple} [cite: #1]}}
\newcommand{\av}[1]{{\color{gray} [AV: #1]}}
\newcommand{\bb}[1]{{\textbf{#1}}}
\newcommand{\uu}[1]{{\underline{#1}}}

\title{End-to-end representation learning for Correlation Filter based tracking}

\author{
Jack Valmadre\thanks{Equal first authorship.}
\quad Luca Bertinetto$^{*}$
\quad Jo\~ao F.\ Henriques
\quad Andrea Vedaldi
\quad Philip H.\ S.\ Torr \\
\vspace{10\baselineskip}
University of Oxford\\
{\tt\small \{name.surname\}@eng.ox.ac.uk}
}

\maketitle
\thispagestyle{empty}

\begin{abstract}
The Correlation Filter is an algorithm that trains a linear template to discriminate between images and their translations.
It is well suited to object tracking because its formulation in the Fourier domain provides a fast solution, enabling the detector to be re-trained once per frame.
Previous works that use the Correlation Filter, however, have adopted features that were either manually designed or trained for a different task.
This work is the first to overcome this limitation by interpreting the Correlation Filter learner, which has a closed-form solution, as a differentiable layer in a deep neural network.
This enables learning deep features that are tightly coupled to the Correlation Filter.
Experiments illustrate that our method has the important practical benefit of allowing lightweight architectures to achieve state-of-the-art performance at high framerates.
\end{abstract}
\section{Introduction}\label{sec:intro}

Deep neural networks are a powerful tool for learning image representations in computer vision applications.
However, training deep networks online, in order to capture previously unseen object classes from one or few examples, is challenging.
This problem emerges naturally in applications such as visual object tracking, where the goal is to re-detect an object over a video with the sole supervision of a bounding box at the beginning of the sequence.
The main challenge is the lack of a-priori knowledge of the target object, which can be of any class.

The simplest approach is to disregard the lack of a-priori knowledge and adapt a pre-trained deep convolutional neural network (CNN) to the target, for example by using stochastic gradient descent (SGD), the workhorse of deep network optimization~\cite{wang2015transferring,nam2016mdnet,zhai2016deep}.
The extremely limited training data and large number of parameters make this a difficult learning problem.
Furthermore, SGD is quite expensive for online adaptation~\cite{wang2015transferring,nam2016mdnet}.

A possible answer to these shortcomings is to have no online adaptation of the network.
Recent works have focused on learning deep embeddings that can be used as universal object descriptors~\cite{bertinetto2016fully,held2016learning,tao2016siamese,leal2016learning,chen2016once}.
These methods use a Siamese CNN, trained offline to discriminate whether two image patches contain the same object or not.
The idea is that a powerful embedding will allow the detection (and thus tracking) of objects via similarity, bypassing the online learning problem.
However, using a fixed metric to compare appearance prevents the learning algorithm from exploiting any video-specific cues that could be helpful for discrimination.

An alternative strategy is to use instead an online learning method such as the \emph{Correlation Filter} (CF).
The CF is an efficient algorithm that learns to discriminate an image patch from the surrounding patches by solving a large ridge regression problem extremely efficiently~\cite{bolme2010visual,henriques2015high}.
It has proved to be highly successful in object tracking (\eg~\cite{danelljan2014accurate,li2014scale,ma2015long,Bertinetto_2016_CVPR}), where its efficiency enables a tracker to adapt its internal model of the object on the fly at every frame.
It owes its speed to a Fourier domain formulation, which allows the ridge regression problem to be solved with only a few applications of the Fast Fourier Transform (FFT) and cheap element-wise operations.
Such a solution is, by design, much more efficient than an iterative solver like SGD, and still allows the discriminator to be tailored to a specific video, contrary to the embedding methods.


The challenge, then, is to combine the online learning efficiency of the CF with the discriminative power of CNN features trained offline.
This has been done in several works (\eg~\cite{ma2015hierarchical,danelljan2015convolutional,danelljan2016beyond,wang2015transferring}), which have shown that CNNs and CFs are complementary and their combination results in improved performance. 

However, in the aforementioned works, the CF is simply applied on top of pre-trained CNN features, without any deep integration of the two methods.
End-to-end training of deep architectures is generally preferable to training individual components separately.
The reason is that in this manner the free parameters in all components can co-adapt and cooperate to achieve a single objective.
Thus it is natural to ask whether a CNN-CF combination can also be trained end-to-end with similar benefits.

The key step in achieving such integration is to interpret the CF as a differentiable CNN layer, so that errors can be propagated through the CF back to the CNN features.
This is challenging, because the CF itself is the solution of a learning problem.
Hence, this requires to differentiate the solution of a large linear system of equations.
This paper provides a closed-form expression for the derivative of the Correlation Filter.
Moreover, we demonstrate the practical utility of our approach in training CNN architectures end-to-end.

We present an extensive investigation into the effect of incorporating the CF into the fully-convolutional Siamese framework of Bertinetto \etal~\cite{bertinetto2016fully}.
We find that the CF does not improve results for networks that are sufficiently deep.
However, our method enables ultra-lightweight networks of a few thousand parameters to achieve state-of-the-art performance on multiple benchmarks while running at high framerates.

Code and results are available online \footnote{\url{www.robots.ox.ac.uk/~luca/cfnet.html}}.

\section{Related work}\label{sec:related-work}

Since the seminal work of Bolme \etal~\cite{bolme2010visual}, the Correlation Filter has enjoyed great popularity within the tracking community.
Notable efforts have been devoted to its improvement, for example by mitigating the effect of periodic boundaries~\cite{fernandez2013zero,kiani2015correlation,danelljan2015learning}, incorporating multi-resolution feature maps~\cite{ma2015hierarchical,danelljan2016beyond} and augmenting the objective with a more robust loss~\cite{rodriguez2013maximum}.
For the sake of simplicity, in this work we adopt the basic formulation of the Correlation Filter.

Recently, several methods based on Siamese networks have been introduced~\cite{tao2016siamese,held2016learning,bertinetto2016fully}, raising interest in the tracking community for their simplicity and competitive performance.
For our method, we prefer to build upon the fully-convolutional Siamese architecture~\cite{bertinetto2016fully}, as it enforces the prior that the appearance similarity function should commute with translation.

At its core, the Correlation Filter layer that we introduce amounts to computing the solution to a regularized deconvolution problem, not to be confused with upsampling convolution layers that are sometimes referred to as ``deconvolution layers''~\cite{long2015fully}.
Before it became apparent that algorithms such as SGD are sufficient for training deep networks, Zeiler \etal~\cite{zeiler2010deconvolutional} introduced a deep architecture in which each layer solves a convolutional sparse coding problem.
In contrast, our problem has a closed-form solution since the Correlation Filter employs quadratic regularization rather than 1-norm regularization.

The idea of back-propagating gradients through the solution to an optimization problem during training has been previously investigated.
Ionescu~\etal~\cite{ionescu2015matrix} and Murray~\cite{murray2016differentiation} have presented back-propagation forms for the SVD and Cholesky decomposition respectively, enabling gradient descent to be applied to a network that computes the solution to either a system of linear equations or an eigenvalue problem.
Our work can be understood as an efficient back-propagation procedure through the solution to a system of linear equations, where the matrix has circulant structure.

When the solution to the optimization problem is obtained iteratively, an alternative is to treat the iterations as a Recurrent Neural Network, and to explicitly unroll a fixed number of iterations~\cite{zheng2015conditional}.
Maclaurin \etal \cite{maclaurin2015gradient} go further and back-propagate gradients through an entire SGD learning procedure, although this is computationally demanding and requires judicious bookkeeping.
Gould \etal~\cite{gould2016differentiating} have recently considered differentiating the solution to general $\arg\min$ problems without restricting themselves to iterative procedures.
However, these methods are unnecessary in the case of the Correlation Filter, as it has a closed-form solution.


Back-propagating through a learning algorithm invites a comparison to meta-learning.
Recent works~\cite{vinyals2016matching,bertinetto2016learning} have proposed feed-forward architectures that can be interpreted as learning algorithms, enabling optimization by gradient descent.
Rather than adopt an abstract definition of learning, this paper propagates gradients through a conventional learning problem that is already widely used.

\section{Method}
\label{sec:method}

We briefly introduce a framework for learning embeddings with Siamese networks (Section~\ref{sec:siam-fc}) and the use of such an embedding for object tracking (Section~\ref{sec:siam-track}) before presenting the \cfnet architecture (Section~\ref{sec:cfnets}). 
We subsequently derive the expressions for evaluation and back-propagation of the main new ingredient in our networks, the Correlation Filter layer, which performs online learning in the forward pass (Section~\ref{sec:cf}).
\subsection{Fully-convolutional Siamese networks}
\label{sec:siam-fc}

Our starting point is a network similar to that of~\cite{bertinetto2016fully}, which we later modify in order to allow the model to be interpreted as a Correlation Filter tracker.
The fully-convolutional Siamese framework considers pairs $(x',z')$ comprising a training image $x'$ and a test image $z'$\footnote{Note that this differs from \cite{bertinetto2016fully}, in which the target object and search area were instead denoted $z$ and $x$ respectively.}.
The image $x'$ represents the object of interest (\eg an image patch centered on the target object in the first video frame), while $z'$ is typically larger and represents the search area (\eg the next video frame).

Both inputs are processed by a CNN $f_\rho$ with learnable parameters $\rho$. This yields two feature maps, which are then cross-correlated:
\begin{equation}\label{eq:siamese-pred}
g_\rho(x', z') = f_\rho(x') \star f_\rho(z') \enspace .
\end{equation}
Eq.~\ref{eq:siamese-pred} amounts to performing an exhaustive search of the pattern $x'$ over the test image $z'$. The goal is for the maximum value of the response map (left-hand side of eq.~\ref{eq:siamese-pred}) to correspond to the target location.

To achieve this goal, the network is trained offline with millions of random pairs $(x_i', z_i')$ taken from a collection of videos.
Each example has a spatial map of labels $c_i$ with values in $\{-1, 1\}$, with the true object location belonging to the positive class and all others to the negative class.
Training proceeds by minimizing an element-wise logistic loss $\ell$ over the training set:
\begin{equation}\label{eq:siamese-train}
\arg\min_{\rho} \quad \sum_i \ell\left(g_\rho(x'_i, z'_i), \, c_i\right) \enspace .
\end{equation}
%



\subsection{Tracking algorithm}\label{sec:siam-track}

The network itself only provides a function to measure the similarity of two image patches. 
To apply this network to object tracking, it is necessary to combine this with a procedure that describes the logic of the tracker.
Similar to~\cite{bertinetto2016fully}, we employ a simplistic tracking algorithm to assess the utility of the similarity function.

Online tracking is performed by simply evaluating the network in forward-mode.
The feature representation of the target object is compared to that of the search region, which is obtained in each new frame by extracting a window centred at the previously estimated position, with an area that is four times the size of the object.
The new position of the object is taken to be the location with the highest score.

The original fully-convolutional Siamese network simply compared every frame to the initial appearance of the object.
In contrast, we compute a new template in each frame and then combine this with the previous template in a moving average.

\subsection{Correlation Filter networks}\label{sec:cfnets}

We propose to modify the baseline Siamese network of eq.~\ref{eq:siamese-pred} with a Correlation Filter block between $x$ and the cross-correlation operator. The resulting architecture is illustrated in Figure~\ref{fig:pipeline}. This change can be formalized as:
\begin{equation}\label{eq:cf-net-pred}
h_{\rho, s, b}(x', z') = s \, \omega\left(f_\rho(x')\right) \star f_\rho(z') + b
\end{equation}
The CF block $w=\omega(x)$ computes a standard CF template $w$ from the training feature map $x = f_{\rho}(x')$ by solving a ridge regression problem in the Fourier domain~\cite{henriques2015high}.
Its effect can be understood as crafting a discriminative template that is robust against translations.
It is necessary to introduce scalar parameters $s$ and $b$ (scale and bias) to make the score range suitable for logistic regression.
Offline training is then performed in the same way as for a Siamese network (Section~\ref{sec:siam-fc}), replacing $g$ with $h$ in eq.~\ref{eq:siamese-train}.

We found that it was important to provide the Correlation Filter with a large region of context in the training image, which is consistent with the findings of Danelljan et al.~\cite{danelljan2015learning} and Kiani et al.~\cite{kiani2015correlation}.
To reduce the effect of circular boundaries, the feature map $x$ is pre-multiplied by a cosine window~\cite{bolme2010visual} and the final template is cropped~\cite{valmadre2014learning}.

Notice that the forward pass of the architecture in Figure~\ref{fig:pipeline} corresponds exactly to the operation of a standard CF tracker~\cite{henriques2015high,danelljan2014accurate,ma2015long,bertinetto2016fully} with CNN features, as proposed in previous work~\cite{ma2015hierarchical,danelljan2015convolutional}. However, these earlier networks were not trained end-to-end. The novelty is to compute the derivative of the CF template with respect to its input so that a network incorporating a CF can be trained end-to-end.
\subsection{Correlation Filter}\label{sec:cf}

We now show how to back-propagate gradients through the Correlation Filter solution efficiently and in closed form via the Fourier domain.

\paragraph{Formulation.}

Given a scalar-valued image $x \in \R^{m \times m}$, the Correlation Filter is the template $w \in \R^{m \times m}$ whose inner product with each circular shift of the image $x * \delta_{-u}$ is as close as possible to a desired response $y[u]$ \cite{henriques2015high}, minimizing
\begin{equation}
\sum_{u \in \mathcal{U}} \left( \langle x * \delta_{-u}, w \rangle - y[u] \right)^{2}
= \| w \star x - y \|^{2} \enspace .
\end{equation}
Here, $\mathcal{U} = \{0, \dots, m-1\}^2$ is the domain of the image, $y \in \R^{m \times m}$ is a signal whose $u$-th element is $y[u]$, and $\delta_{\tau}$ is the translated Dirac delta function $\delta_{\tau}[t] = \delta[t-\tau]$.
In this section, we use $*$ to denote circular convolution and $\star$ to denote circular cross-correlation.
Recall that convolution with the translated $\delta$ function is equivalent to translation $(x * \delta_{\tau})[t] = x[t - \tau \bmod m]$.
Incorporating quadratic regularization to prevent overfitting, the problem is to find
%
\begin{equation}
\arg\min_{w} \quad \frac{1}{2 n} \| w \star x - y \|^{2} + \frac{\lambda}{2} \| w \|^{2}
\label{eq:corr-filt-objective}
\end{equation}
where $n = |\mathcal{U}|$ is the effective number of examples.

The optimal template $w$ must satisfy the system of equations (obtained via the Lagrangian dual, see Appendix~C, supplementary material) 
\begin{equation}
\label{eq:corr-filt-eqs}
\left\{\begin{aligned}
k & = \tfrac{1}{n} (x \star x) + \lambda \delta\\
k * \alpha & = \tfrac{1}{n} y\\
w & = \alpha \star x
\end{aligned}\right.
\end{equation}
%
where $k$ can be interpreted as the signal that defines a circulant linear kernel matrix, and $\alpha$ is a signal comprised of the Lagrange multipliers of a constrained optimization problem that is equivalent to eq.~\ref{eq:corr-filt-objective}.
The solution to eq.~\ref{eq:corr-filt-eqs} can be computed efficiently in the Fourier domain~\cite{henriques2015high},
\begin{subequations}
\label{eq:corr-filter-fourier}
\begin{empheq}[left=\empheqlbrace]{align}
\widehat{k} & = \tfrac{1}{n} (\widehat{x}^{*} \circ \widehat{x}) + \lambda \ones \\
\widehat{\alpha} & = \tfrac{1}{n} \widehat{k}^{-1} \circ \widehat{y} \label{eq:deconv} \\
\widehat{w} & = \widehat{\alpha}^{*} \circ \widehat{x}
\end{empheq}
\end{subequations}
where we use $\widehat{x} = F x$ to denote the Discrete Fourier Transform of a variable, $x^{*}$ to denote the complex conjugate, $\circ$ to denote element-wise multiplication and $\ones$ to denote a signal of ones.
The inverse of element-wise multiplication is element-wise scalar inversion.
Notice that the operations in eq.~\ref{eq:corr-filter-fourier} are more efficiently computed in the Fourier domain, since they involve element-wise operations instead of more expensive convolutions or matrix operators (eq.~\ref{eq:corr-filt-eqs}).
Moreover, the inverse convolution problem (to find $\alpha$ such that $k * \alpha = \frac{1}{n} y$) is the solution to a diagonal system of equations in the Fourier domain (eq.~\ref{eq:deconv}).

\begin{figure}
\centering
\includegraphics[width=0.85\columnwidth]{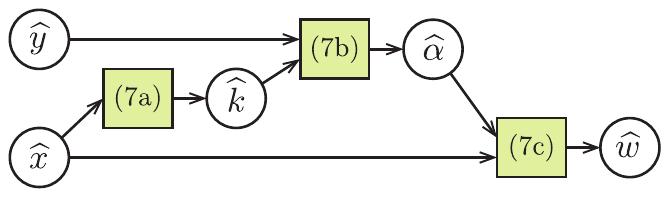}
\caption{Internal computational graph for the Correlation Filter. The boxes denote functions, which are defined in eq.~\ref{eq:corr-filter-fourier}, and the circles denote variables.}
\label{fig:cf-graph}
\end{figure}

\paragraph{Back-propagation.}

We adopt the notation that if $x \in \mathcal{X} = \R^{n}$ is a variable in a computational graph that computes a final scalar loss $\ell \in \R$, then $\nabla_{x} \ell \in \mathcal{X}$ denotes the vector of partial derivatives $(\nabla_{x} \ell)_{i} = \partial \ell / \partial x_{i}$.
If $y \in \mathcal{Y} = \R^{m}$ is another variable in the graph, which is computed directly from $x$ according to $y = f(x)$, then the so-called \emph{back-propagation map} for the function $f$ is a linear map from $\nabla_{y} \ell \in \mathcal{Y}$ to $\nabla_{x} \ell \in \mathcal{X}$.

Appendix~D gives a tutorial review of the mathematical background. In short, the back-propagation map is the linear map which is the adjoint of the differential. 
This property was used by Ionescu~\etal~\cite{ionescu2015matrix} to compute back-propagation maps using matrix differential calculus.
While they used the matrix inner product $\langle X, Y \rangle = \trace(X^{T} Y)$ to find the adjoint, we use Parseval's theorem, which states that the Fourier transform is unitary (except for a scale factor) and therefore preserves inner products $\langle x, y \rangle \propto \langle \widehat{x}, \widehat{y} \rangle$.

To find the linear map for back-propagation through the Correlation Filter, we first take the differentials of the system of equations in eq.~\ref{eq:corr-filt-eqs} that defines the template $w$
\begin{equation}
\left\{\begin{aligned}
dk & = \tfrac{1}{n} (dx \star x + x \star dx) \\
dk * \alpha + k * d\alpha & = \tfrac{1}{n} dy \\
dw & = d\alpha \star x + \alpha \star dx
\end{aligned}\right.
\end{equation}
and then take the Fourier transform of each equation and re-arrange to give the differential of each dependent variable in Figure~\ref{fig:cf-graph} as a linear function (in the Fourier domain) of the differentials of its input variables
\begin{subequations}
\label{eq:fourier-diff}
\begin{empheq}[left=\empheqlbrace]{align}
\label{eq:dk} \widehat{dk} & = \tfrac{1}{n} (\widehat{dx}^{*} \circ \widehat{x} + \widehat{x}^{*} \circ \widehat{dx}) \\
\label{eq:da} \widehat{d\alpha} & = \widehat{k}^{-1} \circ \big[\tfrac{1}{n} \widehat{dy} - \widehat{dk} \circ \widehat{\alpha}\big] \\
\label{eq:dw} \widehat{dw} & = \widehat{d\alpha}^{*} \circ \widehat{x} + \widehat{\alpha}^{*} \circ \widehat{dx} \enspace .
\end{empheq}
\end{subequations}
Note that while these are complex equations, that is simply because they are the Fourier transforms of real equations.
The derivatives themselves are all computed with respect to real variables.

The adjoints of these linear maps define the overall back-propagation map from $\nabla_{w} \ell$ to $\nabla_{x} \ell$ and $\nabla_{y} \ell$. We defer the derivation to Appendix~\ref{app:corr-filt-back-prop} and present here the final result,
\begin{equation}
\left\{\begin{aligned}
\widehat{\nabla_{\alpha} \ell} & = \widehat{x} \circ (\widehat{\nabla_{w} \ell})^{*} \\
\widehat{\nabla_{y} \ell} & = \tfrac{1}{n} \widehat{k}^{-*} \circ \widehat{\nabla_{\alpha} \ell} \\
\widehat{\nabla_{k} \ell} & = -\widehat{k}^{-*} \circ \widehat{\alpha}^{*} \circ \widehat{\nabla_{\alpha} \ell} \\
\widehat{\nabla_{x} \ell} & = \widehat{\alpha} \circ \widehat{\nabla_{w} \ell} + \tfrac{2}{n} \widehat{x} \circ \Re\{\widehat{\nabla_{k} \ell}\} \enspace .
\end{aligned}\right.
\label{eq:back-prop-result}
\end{equation}
It is necessary to compute forward Fourier transforms at the start and inverse transforms at the end. The extension to multi-channel images is trivial and given in Appendix~E (supplementary material). 

As an interesting aside, we remark that, since we have the gradient of the loss with respect to the ``desired'' response $y$, it is actually possible to optimize for this parameter rather than specify it manually.
However, in practice we did not find learning this parameter to improve the tracking accuracy compared to the conventional choice of a fixed Gaussian response~\cite{bolme2010visual,henriques2015high}.


\section{Experiments}\label{sec:experiments}
The principal aim of our experiments is to investigate the effect of incorporating the Correlation Filter during training.
We first compare against the symmetric Siamese architecture of Bertinetto \etal~\cite{bertinetto2016fully}.
We then compare the end-to-end trained \cfnet to a variant where the features are replaced with features that were trained for a different task.
Finally, we demonstrate that our method achieves state-of-the-art results.

\subsection{Evaluation criteria}
Popular tracking benchmarks like VOT~\cite{Kristan2016a} and OTB~\cite{wu2013online,wu2015object} have made all ground truth annotations available and do not enforce a validation/test split.
However, in order to avoid overfitting to the test set in design choices and hyperparameter selection, we consider OTB-2013, OTB-50 and OTB-100 as our \emph{test set} and 129 videos from VOT-2014, VOT-2016 and Temple-Color~\cite{liang2015encoding} as our \emph{validation set}, excluding any videos which were already assigned to the test set.
We perform all of our tracking experiments in Sections~\ref{sec:baseline-comparison}, \ref{sec:transfer} and~\ref{sec:adaptation} on the validation set with the same set of ``natural'' hyperparameters, which are reasonable for all methods and not tuned for any particular method.

As in the OTB benchmark~\cite{wu2013online,wu2015object}, we quantify the performance of the tracker on a sequence in terms of the average overlap (intersection over union) of the predicted and ground truth rectangles in all frames.
The success rate of a tracker at a given threshold $\tau$ corresponds to the fraction of frames in which the overlap with the ground truth is at least $\tau$.
This is computed for a uniform range of 100 thresholds between 0 and 1, effectively constructing the cumulative distribution function.
Trackers are compared using the area under this curve.

Mimicking the TRE (Temporal Robustness Evaluation) mode of OTB, we choose three equispaced points per sequence and run the tracker from each until the end.
Differently from the OTB evaluation, when the target is \emph{lost} (\ie the overlap with the ground truth becomes zero) the tracker is terminated and an overlap of zero is reported for all remaining frames.

Despite the large number of videos, we still find that the performance of similarity networks varies considerably as training progresses.
To mitigate this effect, we average the final tracking results that are obtained using the parameters of the network at epochs 55, 60, \dots, 95, 100 (the final epoch) to reduce the variance.
These ten results are used to estimate the standard deviation of the distribution of results, providing error bars for most figures in this section.
While it would be preferable to train all networks to convergence multiple times with different
random seeds, this would require significantly more resources.

\subsection{Comparison to Siamese baseline}
\label{sec:baseline-comparison}


\begin{figure}
\centering
\includegraphics{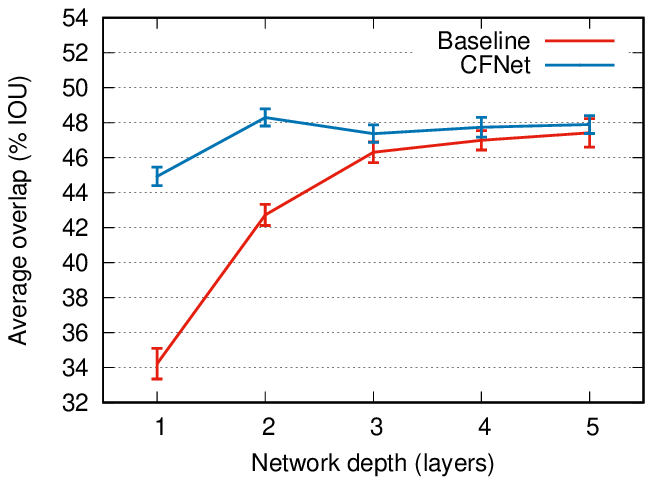}
\caption{Tracker accuracy for different network depths, on the 129 videos of the validation set. Error bars indicate two standard deviations. Refer to section~\ref{sec:baseline-comparison} for more details.
All figures best viewed in colour.}
\label{fig:main_expm}
\end{figure}

\begin{figure*}
    \centering
    \begin{subfigure}[t]{0.32\textwidth}
        \centering
        \includegraphics[width=\textwidth]{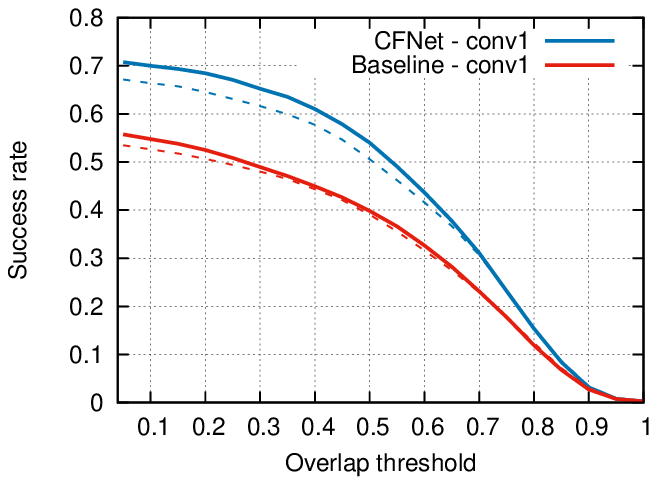}
    \end{subfigure}
    \begin{subfigure}[t]{0.32\textwidth}
        \centering
        \includegraphics[width=\linewidth]{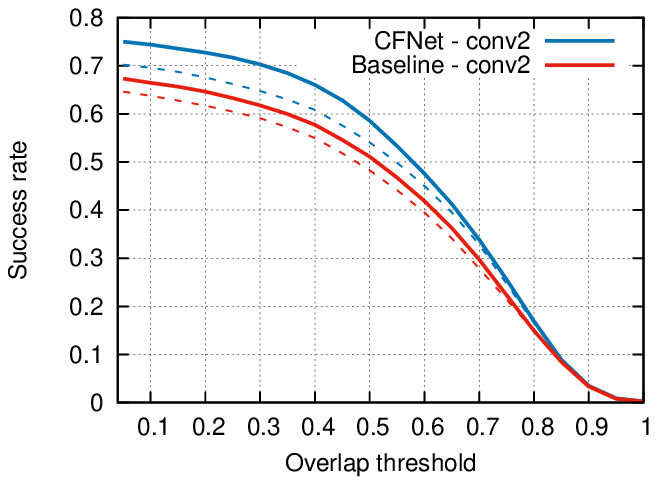}
    \end{subfigure}
    \begin{subfigure}[t]{0.32\textwidth}
        \centering
        \includegraphics[width=\linewidth]{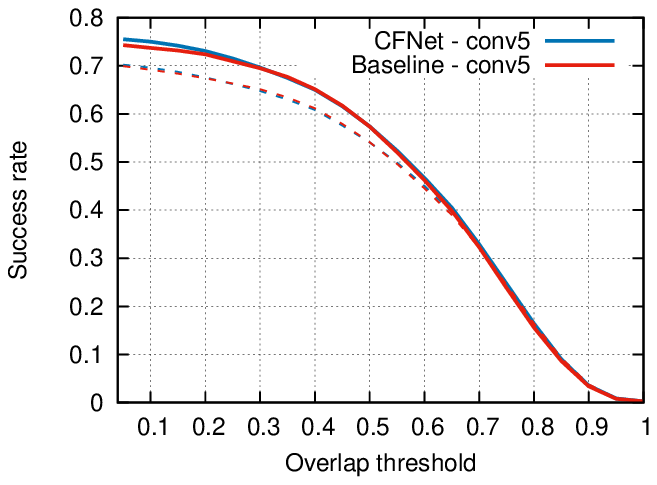}
    \end{subfigure}
    \caption{Success rates of rectangle overlap for individual trackers on the validation set.
    Solid and dotted lines represent methods that update the template with a running average learning rate of 0.01 and 0, respectively.}
    \label{fig:overlap-cdf}
\end{figure*}

Figures~\ref{fig:main_expm} and~\ref{fig:overlap-cdf} compare the accuracy of both methods on the validation set for networks of varying depth.
The feature extraction network of depth $n$ is terminated after the $n$-th linear layer, including the following ReLU but not the following pooling layer (if any).

Our baseline diverges slightly from~\cite{bertinetto2016fully} in two ways.
Firstly, we reduce the total stride of the network from 8 to 4 (2 at conv1, 2 at pool1) to avoid training Correlation Filters with small feature maps.
Secondly, we always restrict the final layer to 32 output channels in order to preserve the high speed of the method with larger feature maps.
These changes did not have a negative effect on the tracking performance of SiamFC.

The results show that \cfnet is significantly better than the baseline when shallow networks are used to compute features.
Specifically, it brings a relative improvement of 31\% and 13\% for networks of depth one and two respectively.
At depths three, four and five, the difference is much less meaningful.
\cfnet is relatively unaffected by the depth of the network, whereas the performance of the baseline increases steadily and significantly with depth.
It seems that the ability of the Correlation Filter to adapt the distance metric to the content of the training image is less important given a sufficiently expressive embedding function.

The CF layer can be understood to encode prior knowledge of the test-time procedure.
This prior may become redundant or even overly restrictive when enough model capacity and data are available.
We believe this explains the saturation of \cfnet performance when more than two convolutional layers are used.

Figure~\ref{fig:overlap-cdf} additionally shows that updating the template is always helpful, for both Baseline and \cfnet architectures, at any depth.





\subsection{Feature transfer experiment}
\label{sec:transfer}


\begin{figure}
\centering
\includegraphics{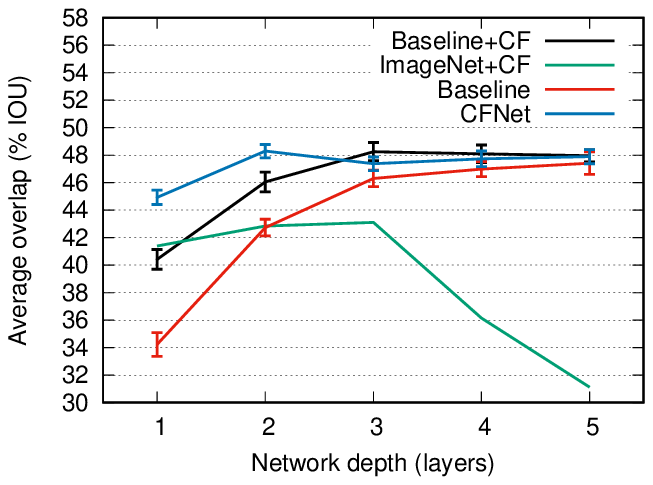}
\caption{
Accuracy of a Correlation Filter tracker when using features obtained via different methods.
Error bars indicate two standard deviations.
Refer to Section~\ref{sec:transfer} for details.
}
\label{fig:transfer}
\end{figure}

The motivation for this work was the hypothesis that incorporating the CF during training will result in features that are better suited to tracking with a CF.
We now compare our end-to-end trained \cfnet to variants that use features from alternative sources: \emph{Baseline+CF} and \emph{ImageNet+CF}.
The results are presented in Figure~\ref{fig:transfer}.

To obtain the curve \mbox{\emph{Baseline+CF}} we trained a baseline Siamese network of the desired depth and then combined those features with a CF during tracking.
Results show that taking the CF into account during offline training is critical at depth one and two.
However, it seems redundant when more convolutional layers are added, since using features from the \emph{Baseline} in conjunction with the CF achieves similar performance.



The \emph{ImageNet+CF} variant employs features taken from a network trained to solve the ImageNet classification challenge~\cite{ILSVRC15}.
The results show that these features, which are often the first choice for combining CFs with CNNs~\cite{danelljan2015convolutional,danelljan2016beyond,ma2015hierarchical,nam2016mdnet,wang2015transferring,zhai2016deep}, are significantly worse than those learned by \emph{CFNet} and the \emph{Baseline} experiment.
The particularly poor performance of these features at deeper layers is somewhat unsurprising, since these layers are expected to have greater invariance to position when trained for classification.



\subsection{Importance of adaptation}
\label{sec:adaptation}

\begin{figure}
\centering
\includegraphics{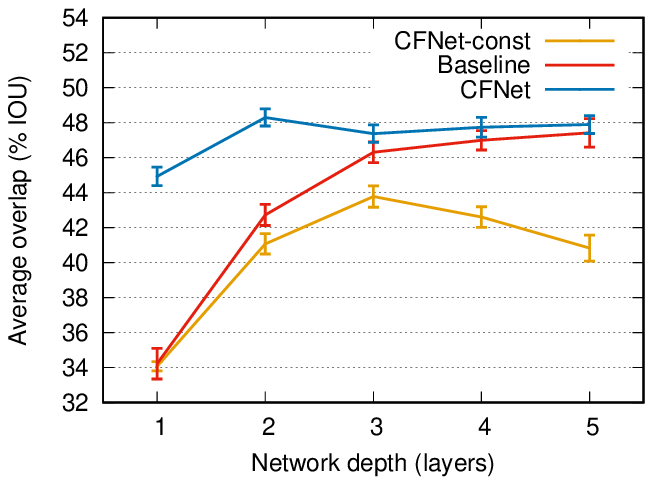}
\caption{
  Comparison of \cfnet to a ``constant'' variant of the architecture, in which the Lagrange multipliers do not depend on the image (section~\ref{sec:adaptation}).
  Error bars indicate two standard deviations.
 }
\label{fig:cf-const}
\end{figure}

\begin{table*}
\centering
\footnotesize
\begin{tabular}{l c c    c c c c c    c c c c c    c c c c}
\toprule
 & & & \multicolumn{4}{c}{\textbf{OTB-2013}} & & \multicolumn{4}{c}{\textbf{OTB-50}} & & \multicolumn{4}{c}{\textbf{OTB-100}}\\ [0.5ex]
 & & & \multicolumn{2}{c}{\textbf{OPE}} & \multicolumn{2}{c}{\textbf{TRE}} & & \multicolumn{2}{c}{\textbf{OPE}} & \multicolumn{2}{c}{\textbf{TRE}} & & \multicolumn{2}{c}{\textbf{OPE}} & \multicolumn{2}{c}{\textbf{TRE}}\\ [0.5ex]
 \textbf{Method} & \textbf{speed} (fps.) & & IoU & prec. & IoU & prec. & & IoU & prec. & IoU & prec. & & IoU & prec. & IoU & prec.\\ [0.5ex]
\midrule
CFNet-conv1 & 83 & & 							57.8 	  & 77.6 	  & 58.6 	  & 77.6 & & 			48.8 	  & 65.3 	  & 51.0 	  & 67.9 & & 		53.6 	  & 71.3 	  & 55.9 	  & 72.6\\
CFNet-conv2 & 75 & & 							61.1 	  & 80.7 	  & \bb{64.0} & \bb{84.8} & & 		53.0  	  & 70.2 	  & 56.5 	  & 75.3 & & 		56.8 	  & 74.8 	  & 60.6 	  & 79.1\\
Baseline+CF-conv3 & 67 & & 						61.0 	  & \uu{82.2} & \uu{63.1} & \uu{83.9} & & 		\uu{53.8} & \uu{72.3} & \bb{57.4} & \bb{76.7} & & 	\bb{58.9} & \uu{77.7} & \uu{61.1} & \bb{79.8}\\
CFNet-conv5 & 43 & & 							61.1 	  & 80.3 	  & 62.6 	  & 82.5 & & 			\bb{53.9} & \bb{73.2} & \uu{56.6} & \uu{75.9} & & 	58.6 	  & \uu{77.7} & 60.8 	  & 78.8\\
Baseline-conv5 & 52 & & 						\bb{61.8} & 80.6      & \bb{64.0} & 83.7 & & 			51.7 	  & 68.3 	  & 56.1 	  & 74.2 & & 		\uu{58.8} & 76.9 	  & \bb{61.6} & \uu{79.7}\\
\midrule
SiamFC-3s~\cite{bertinetto2016fully} & & & 	60.7 	  & 81.0 	  & 61.8 	  & 82.2 & & 			51.6 	  & 69.2 	  & 55.5 	  & 75.2 & & 			58.2 	  & 77.0 	  & 60.5 	  & 79.5\\
Staple~\cite{Bertinetto_2016_CVPR} & & & 	60.0 	  & 79.3 	  & 61.7 	  & 80.3 & & 			50.9 	  & 68.1 	  & 54.1 	  & 72.6 & & 			58.1 	  & \bb{78.4} & 60.4 	  & 78.9\\
LCT~\cite{ma2015long} & & & 				\uu{61.2} & \bb{86.2} & 59.4 	  & 81.3 & & 			49.2 	  & 69.1 	  & 49.5 	  & 67.4 & & 			56.2 	  & 76.2 	  & 56.9 	  & 74.5\\
SAMF~\cite{li2014scale} & & & 				--   	  & --   	  & --   	  & --   & & 			46.2 	  & 63.9 	  & 51.4 	  & 70.9 & & 			53.9 	  & 74.6 	  & 57.7 	  & 77.6\\
DSST~\cite{danelljan2014accurate} & & & 	55.4 	  & 74.0 	  & 56.6 	  & 73.8 & & 			45.2 	  & 60.4 	  & 48.4 	  & 64.1 & & 			51.3 	  & 68.0 	  & --   	  & --  \\
\bottomrule
\end{tabular}
\caption{Perfomance as overlap (IoU) and precision produced by the OTB toolkit for the OTB-2013, OTB-50 and OTB-100 datasets.
The \textbf{first} and \underline{second} best results are highlighted in each column.
For details refer to Section~\ref{sec:otb}.}
\label{tab:otb}
\end{table*}

For a multi-channel CF, each channel~$p$ of the template~$w$ can be obtained as $w_{p} = \alpha \star x_{p}$, where $\alpha$ is itself a function of the exemplar $x$ (Appendix~C, supplementary material).
To verify the importance of the online adaptation that solving a ridge regression problem at test time should provide, we propose a ``constant'' version of the Correlation Filter (\mbox{\emph{CFNet-const}}) where the vector of Lagrange multipliers $\alpha$ is instead a parameter of the network that is learned offline and remains fixed at test time.

Figure~\ref{fig:cf-const} compares \cfnet to its constant variant.
\cfnet is consistently better, demonstrating that in order to improve over the baseline Siamese network it is paramount to back-propagate through the solution to the inverse convolution problem that defines the Lagrange multipliers.


\subsection{Comparison with the state-of-the-art}\label{sec:otb}
We use the OTB-2013/50/100 benchmarks to confirm that our results are on par with the state-of-the-art.
All numbers in this section are obtained using the OTB toolkit~\cite{wu2013online}.
We report the results for the three best instantiations of \cfnet from Figure~\ref{fig:transfer} (\emph{\cfnet-conv2}, \emph{\cfnet-conv5}, \emph{Baseline+CF-conv3}), the best variant of the baseline (\emph{Baseline-conv5}) and the most promising single-layer network (\emph{\cfnet-conv1}).
We compare our methods against state-of-the-art trackers that can operate in real-time: SiamFC-3s~\cite{bertinetto2016fully}, Staple~\cite{Bertinetto_2016_CVPR} and LCT~\cite{ma2015long}.
We also include the recent SAMF~\cite{li2014scale} and DSST~\cite{danelljan2014accurate} for reference.

For the evaluation of this section, we use a different set of tracking hyperparameters per architecture, chosen to maximize the performance on the validation set after a random search of 300 iterations.
More details are provided in the supplementary material.
For the few greyscale sequences present in OTB, we re-train each architecture using exclusively greyscale images.

Both overlap (IoU) and precision scores~\cite{wu2015object} are reported for OPE (one pass) and TRE (temporal robustness) evaluations.
For OPE, the tracker is simply run once on each sequence, from the start to the end.
For TRE, the tracker is instead started from twenty different starting points, and run until the end from each.
We observed that this ensures more robust and reliable results compared to OPE.

Similarly to the analysis on the validation set, \emph{\cfnet-conv2} is among the top performers and its accuracy rivals that of \emph{Baseline-conv5}, which possesses approximately 30$\times$ as many parameters.
In general, our best proposed \cfnet variants are superior (albeit modestly) to the state-of-the-art.
In order to focus on the impact of our contribution, we decided to avoid including orthogonal improvements which can often be found in the tracking literature (\eg bounding box regression~\cite{nam2016mdnet}, ensembling of multiple cues~\cite{ma2015long,Bertinetto_2016_CVPR}, optical flow~\cite{tao2016siamese}).

\subsection{Speed and practical benefits}\label{sec:speed}

\begin{figure}
\centering
\includegraphics[width=\columnwidth]{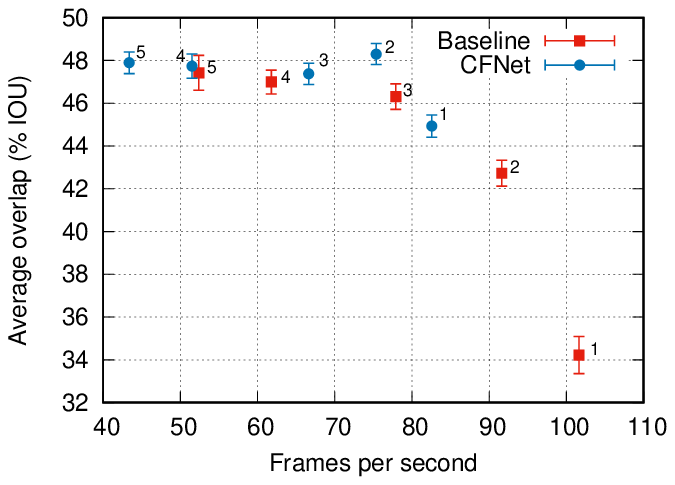}
\caption{
Tracker accuracy versus speed for \cfnet and Siamese baseline.
Labels indicate network depth.
\cfnet enables better accuracy to be obtained at higher speeds using shallower networks.
Error bars indicate two standard deviations.
Refer to section~\ref{sec:speed} for details.}
\label{fig:speed}
\end{figure}

The previous sections have demonstrated that there is a clear benefit to integrating Correlation Filters into Siamese networks when the feature extraction network is relatively shallow.
Shallow networks are practically advantageous in that they require fewer operations and less memory to evaluate and store.
To understand the trade-off, Figure~\ref{fig:speed} reports the speed and accuracy of both \cfnet and the baseline for varying network depth\footnote{The speed was measured using a 4.0GHz Intel i7 CPU and an NVIDIA Titan X GPU.}.

This plot suggests that the two-layer \cfnet could be the most interesting variant for practitioners requiring an accurate tracking algorithm that operates at high framerates.
It runs at 75 frames per second and has less than 4\% of the parameters of the five-layer baseline, requiring only 600kB to store.
This may be of particular interest for embedded devices with limited memory.
In contrast, methods like DeepSRDCF~\cite{danelljan2015convolutional} and C-COT~\cite{danelljan2016beyond}, which use out-of-the-box deep features for the Correlation Filter, run orders of magnitude slower.
Even the one-layer \cfnet remains competitive despite having less than 1\% of the parameters of the five-layer baseline and requiring under 100kB to store.

\section{Conclusion}\label{sec:conclusion}

This work proposes the Correlation Filter network, an asymmetric architecture that back-propagates gradients through an online learning algorithm to optimize the underlying feature representation.
This is made feasible by establishing an efficient back-propagation map for the solution to a system of circulant equations.

Our empirical investigation reveals that, for a sufficiently deep Siamese network, adding a Correlation Filter layer does not significantly improve the tracking accuracy.
We believe this is testament to the power of deep learning given sufficient training data.
However, incorporating the Correlation Filter into a similarity network during training does enable shallow networks to rival their slower, deeper counterparts.

Future research may include extensions to account for adaptation over time, and back-propagating gradients through learning problems for related tasks such as one-shot learning and domain adaptation.


\appendix

\section{Implementation details}

We follow the procedure of~\cite{bertinetto2016fully} to minimize the loss (equation~\ref{eq:siamese-train}) through SGD, with the Xavier-improved parameters initialization and using mini-batches of size 8.
We use all the 3862 training videos of ImageNet Video~\cite{ILSVRC15}, containing more than 1 million annotated frames, with multiple objects per frame.
Training is conducted for 100 epochs, each sampling approximately 12 pairs $(x_i', z_i')$ from each video, randomly extracted so that they are at most 100 frames apart.

During tracking, a spatial cosine window is multiplied with the score map to penalize large displacements.
Tracking in scale space is achieved by evaluating the network at the scale of the previous object and at one adjacent scale on either side, with a geometric step of 1.04.
Updating the scale is discouraged by multiplying the responses of the scaled object by 0.97.
To avoid abrupt transitions of object size, scale is updated using a rolling average with learning rate 0.6.
%

\section{Back-propagation for the Correlation Filter}
\label{app:corr-filt-back-prop}

As described in Appendix~D (supplementary material), the back-propagation map is the adjoint of the linear maps that is the differential.
These linear maps for the Correlation Filter are presented in eq.~\ref{eq:fourier-diff}.
We are free to obtain these adjoint maps in the Fourier domain since Parseval's theorem provides the preservation of inner products.
Let $J_{1}$ denote the map $dx \mapsto dk$ in eq.~\ref{eq:dk}.
Hence manipulation of the inner product
\begin{align}
\langle F dk, F J_{1}(dx) \rangle
& = \left\langle \widehat{dk}, \tfrac{1}{n} (\widehat{dx}^{*} \circ \widehat{x} + \widehat{x}^{*} \circ \widehat{dx}) \right\rangle \nonumber \\
& = \tfrac{1}{n} \left[\langle \widehat{dx}, \widehat{dk}^{*} \circ \widehat{x} \rangle
  + \langle \widehat{dk} \circ \widehat{x}, \widehat{dx} \rangle\right] \nonumber \\
& = \left\langle \widehat{dx}, \tfrac{2}{n} \Re\{\widehat{dk}\} \circ \widehat{x} \right\rangle
\end{align}
gives the back-propagation map
\begin{equation}
\widehat{\nabla_{x} \ell} = \tfrac{2}{n} \widehat{x} \circ \Re\{\widehat{\nabla_{k} \ell}\} \enspace .
\end{equation}
Similarly, for the linear map $dk, dy \mapsto d\alpha$ in eq.~\ref{eq:da},
\begin{align}
& \langle F d\alpha, F J_{2}(dk, dy)\rangle
= \left\langle \widehat{d\alpha}, \widehat{k}^{-1} [\tfrac{1}{n} \widehat{dy} - \widehat{dk} \circ \widehat{\alpha}] \right\rangle \nonumber \\
& \; = \left\langle \tfrac{1}{n} \widehat{k}^{-*} \circ \widehat{d\alpha}, \widehat{dy} \right\rangle
  + \left\langle -\widehat{k}^{-*} \circ \widehat{\alpha}^{*} \circ \widehat{d\alpha}, \widehat{dk} \right\rangle
\enspace ,
\end{align}
the back-propagation maps are
\begin{align}
\widehat{\nabla_{y} \ell} & = \tfrac{1}{n} \widehat{k}^{-*} \circ \widehat{\nabla_{\alpha} \ell} \\
\widehat{\nabla_{k} \ell} & = -\widehat{k}^{-*} \circ \widehat{\alpha}^{*} \circ \widehat{\nabla_{\alpha} \ell} \enspace ,
\end{align}
and for the linear map $dx, d\alpha \mapsto dw$ in eq.~\ref{eq:dw},
\begin{align}
& \langle F dw, F J_{3}(dx, d\alpha) \rangle = \langle \widehat{dw}, \widehat{d\alpha}^{*} \circ \widehat{x} + \widehat{\alpha}^{*} \circ \widehat{dx} \rangle \nonumber \\
& \; = \langle \widehat{d\alpha}, \widehat{dw}^{*} \circ \widehat{x} \rangle
  + \langle \widehat{dw} \circ \widehat{\alpha}, \widehat{dx} \rangle
\enspace ,
\end{align}
the back-propagation maps are
\begin{align}
\widehat{\nabla_{\alpha} \ell} & = \widehat{x} \circ (\widehat{\nabla_{w} \ell})^{*} \enspace , \\
\widehat{\nabla_{x} \ell} & = \widehat{\alpha} \circ \widehat{\nabla_{w} \ell} \enspace .
\end{align}
The two expressions for $\nabla_{x} \ell$ above are combined to give the back-propagation map for the entire Correlation Filter block in eq.~\ref{eq:back-prop-result}.


\section{Correlation Filter formulation}
\label{app:corr-filt-soln}

\subsection{Kernel linear regression}

First, consider the general linear regression problem of learning the weight vector $w$ that best maps each of $n$ example input vectors $x_{i} \in \R^{d}$ to their target $y_{i} \in \R$.
The squared error can be expressed
\begin{equation}
\frac{1}{2 n} \sum_{i = 1}^{n} (x_{i}^{T} w - y_{i})^{2}
= \frac{1}{2 n} \|X^{T} w - y\|^{2}
\end{equation}
where $X \in \R^{d \times n}$ is a matrix whose columns are the example vectors and $y \in \R^{n}$ is a vector of the targets.
Incorporating regularization, the problem is
\begin{equation}
\arg \min_{w} \quad \tfrac{1}{2 n} \| X^T w - y \|^{2} + \tfrac{\lambda}{2} \| w \|^{2} \enspace .
\end{equation}
Kernel linear regression can be developed by writing this as a constrained optimization problem
\begin{equation}
\begin{aligned}
&& \arg \min_{w, r} &&& \tfrac{1}{2 n} \|r\|^{2} + \tfrac{\lambda}{2} \|w\|^{2} \\
&& \text{subject to} &&& r = X^T w - y
\end{aligned}
\end{equation}
and then finding a saddle point of the Lagrangian
\begin{equation}
L(w, r, \upsilon) = \tfrac{1}{2 n} \|r\|^{2} + \tfrac{\lambda}{2} \|w\|^{2} + \upsilon^{T}(r - X^T w + y) \enspace .
\end{equation}
The final solution can be obtained from the dual variable
\begin{equation}
w = \tfrac{1}{\lambda} X \upsilon
\end{equation}
and the solution to the dual problem is
\begin{equation}
\upsilon = \tfrac{\lambda}{n} K^{-1} y
\end{equation}
where $K = \frac{1}{n} X^T X + \lambda I$ is the regularized kernel matrix.
It is standard to introduce a scaled dual variable $\alpha = \frac{1}{\lambda} v$ that defines $w$ as a weighted combination of examples
\begin{equation}
w = X \alpha = \sum_{i = 1}^{n} \alpha_{i} x_{i}
\quad \text{ with } \quad
\alpha = \frac{1}{n} K^{-1} y \enspace .
\end{equation}
The kernel matrix is $n \times n$ and therefore the dual solution is more efficient than the primal solution, which requires inversion of a $d \times d$ matrix, when the number of features $d$ exceeds the number of examples~$n$.

\subsection{Single-channel Correlation Filter}

Given a scalar-valued example signal $x$ with domain $\mathcal{U}$ and corresponding target signal $y$, the Correlation Filter $w$ is the scalar-valued signal
\begin{equation}
\arg \min_{w} \quad \frac{1}{2 n} \| w \star x - y \|^{2} + \frac{\lambda}{2} \| w \|^{2}
\end{equation}
where signals are treated as vectors in $\R^{\mathcal{U}}$ and the circular cross-correlation of two signals $w \star x$ is defined
\begin{equation}
(w \star x)[u] = \sum_{t \in \mathcal{U}} w[t] x[u+t \bmod m] \quad \forall u \in \mathcal{U} \enspace .
\end{equation}
The solution from the previous section can then be used by defining $X$ to be the matrix in $\R^{\mathcal{U} \times \mathcal{U}}$ such that $X^{T} w = w \star x$.
It follows that the kernel matrix $K$ belongs to $\R^{\mathcal{U} \times \mathcal{U}}$ and the dual variable $\alpha$ is a signal in $\R^{\mathcal{U}}$.

The key to the correlation filter is that the circulant structure of $X$ enables the solution to be computed efficiently in the Fourier domain.
The matrix $X$ has elements $X[u, t] = x[u + t \bmod m]$.
Since the matrix $X$ is symmetric, the template $w$ is obtained as cross-correlation
\begin{equation}
w = X \alpha = \alpha \star x \enspace .
\end{equation}
The linear map defined by the kernel matrix $K$ is equivalent to convolution with a signal $k$
\begin{equation}
K z = k * z \quad \forall z
\end{equation}
which is defined $k = \frac{1}{n} x \star x + \lambda \delta$, since
\begin{align}
\forall z: F X^T X z
& = F((z \star x) \star x) \nonumber \\
& = \widehat{z} \circ \widehat{x}^{*} \circ \widehat{x}
  = F(z * (x \star x)) \enspace .
\end{align}
Therefore the solution is defined by the equations
\begin{equation}
\left\{\begin{aligned}
k & = \tfrac{1}{n} x \star x + \lambda \delta \\
k * \alpha & = \tfrac{1}{n} y \\
w & = \alpha \star x
\end{aligned}\right.
\end{equation}
and the template can be computed efficiently in the Fourier domain
\begin{equation}
\left\{\begin{aligned}
\widehat{k} & = \tfrac{1}{n} \widehat{x}^{*} \circ \widehat{x} + \lambda \ones \\
\widehat{\alpha} & = \tfrac{1}{n} \widehat{k}^{-1} \circ \widehat{y} \\
\widehat{w} & = \widehat{\alpha}^{*} \circ \widehat{x} \enspace .
\end{aligned}\right.
\end{equation}

\subsection{Multi-channel Correlation Filter}
\label{app:corr-filt-soln-multi}

There is little advantage to the dual solution when training a single-channel Correlation Filter from the circular shifts of a single base example.
However, the dual formulation is much more efficient in the multi-channel case~\cite{henriques2015high}.

For signals with $k$ channels, each multi-channel signal is a collection of scalar-valued signals $x = (x_{1}, \dots, x_{k})$, and the data term becomes
\begin{equation}
\| {\textstyle \sum_{p} w_{p} \star x_{p}} - y \|^{2}
= \| {\textstyle \sum_{p} X_{p}^{T} w_{p}} - y \|^{2}
\end{equation}
and each channel of the template is obtained from the dual variables
\begin{equation}
w_{p} = X_{p} \alpha = \alpha \star x_{p}
\end{equation}
The solution to the dual problem is still $\alpha = \frac{1}{n} K^{-1} y$, however the kernel matrix is now given
\begin{equation}
K = \tfrac{1}{n} {\textstyle \sum_{p} X_{p}^{T} X_{p}} + \lambda I
\end{equation}
and the linear map defined by this matrix is equivalent to convolution with the signal
\begin{equation}
k = \tfrac{1}{n} {\textstyle \sum_{p} x_{p} \star x_{p}} + \lambda \delta \enspace .
\end{equation}

Therefore the solution is defined by the equations
\begin{equation}
\left\{\begin{aligned}
k & = \tfrac{1}{n} {\textstyle \sum_{p} x_{p} \star x_{p}} + \lambda \delta \\
k * \alpha & = \tfrac{1}{n} y \\
w_{p} & = \alpha \star x_{p} \quad \forall p
\end{aligned}\right.
\label{eq:mccf}
\end{equation}
and the template can be computed efficiently in the Fourier domain
\begin{equation}
\left\{\begin{aligned}
\widehat{k} & = \tfrac{1}{n} {\textstyle \sum_{p} \widehat{x}_{p}^{*} \circ \widehat{x}_{p}} + \lambda \ones \\
\widehat{\alpha} & = \tfrac{1}{n} \widehat{k}^{-1} \circ \widehat{y} \\
\widehat{w}_{p} & = \widehat{\alpha}^{*} \circ \widehat{x}_{p} \quad \forall p \enspace .
\end{aligned}\right.
\end{equation}
It is critical that the computation scales only linearly with the number of channels.

\section{Adjoint of the differential}
\label{app:back-prop-adjoint}

Consider a computational graph that computes a scalar loss $\ell \in \R$.
Within this network, consider an intermediate function that computes $y = f(x)$ where $x \in \mathcal{X} = \R^{m}$ and $y \in \mathcal{Y} = \R^{n}$.
Back-propagation computes the gradient with respect to the input $\nabla_{x} \ell \in \mathcal{X}$ from the gradient with respect to the output $\nabla_{y} \ell \in \mathcal{Y}$.

The derivative $\partial f(x)/\partial x$ is a matrix in $\R^{n \times m}$ whose $i j$-th element is the partial derivative $\partial f_{i}(x) / \partial x_{j}$.
This matrix relates the gradients according to
\begin{equation}
(\nabla_{x} \ell)^{T}
= \frac{\partial \ell}{\partial x}
= \frac{\partial \ell}{\partial y} \frac{\partial y}{\partial x}
= (\nabla_{y} \ell)^{T} \frac{\partial f(x)}{\partial x}
\end{equation}
From this it is evident that the back-propagation map is the linear map which is the adjoint of that defined by the derivative.
That is, if the derivative defines the linear map
\begin{equation}
J(u) = \frac{\partial f(x)}{\partial x} u
\end{equation}
then the back-propagation map is the unique linear map $J^{*}$ that satisfies
\begin{equation}
\langle J^{*}(v), u \rangle = \langle v, J(u) \rangle \quad \forall u \in \mathcal{X}, v \in \mathcal{Y}
\end{equation}
and the gradient with respect to the input is obtained $\nabla_{x} \ell = J^{*}(\nabla_{y} \ell)$.
This is the core of reverse-mode differentiation~\cite{griewank2008evaluating}.

An alternative way to obtain the linear map defined by the derivative is to use differential calculus.
Whereas the \emph{derivative} represents this linear map as a matrix with respect to the standard bases, the \emph{differential} represents the linear map as an expression $df(x; dx)$.
This is valuable for working with variables that possess more interesting structure than simple vectors.
This technique has previously been used for matrix structured back-propagation~\cite{ionescu2015matrix}.
In this paper, we use it for circulant structured back-propagation.


\section{Back-propagation for multi-channel case}
\label{app:corr-filt-back-prop-multi}

The differentials of the equations that define the multi-channel CF in eq.~\ref{eq:mccf} are
\begin{equation}
\left\{\begin{aligned}
dk & = \textstyle \tfrac{1}{n} \sum_{p} (dx_{p} \star x_{p} + x_{p} \star dx_{p}) \\
dk * \alpha + k * d\alpha & = \tfrac{1}{n} dy \\
dw_{p} & = d\alpha \star x_{p} + \alpha \star dx_{p} \quad \forall p \enspace ,
\end{aligned}\right.
\end{equation}
and taking the Fourier transforms of these equations gives
\begin{equation}
\left\{\begin{aligned}
\widehat{dk} & = \textstyle \tfrac{1}{n} \sum_{p} \big(\widehat{dx}_{p}^{*} \circ \widehat{x}_{p} + \widehat{x}_{p}^{*} \circ \widehat{dx}_{p}\big) \\
\widehat{d\alpha} & = \widehat{k}^{-1} \circ \big[\tfrac{1}{n} \widehat{dy} - \widehat{dk} \circ \widehat{\alpha}\big] \\
\widehat{dw}_{p} & = \widehat{d\alpha}^{*} \circ \widehat{x}_{p} + \widehat{\alpha}^{*} \circ \widehat{dx}_{p} \quad \forall p \enspace .
\end{aligned}\right.
\end{equation}

Now, to find the adjoint of the map $dx \mapsto dk$, we re-arrange the inner product
\begin{align}
& \langle F dk, F J_{1}(dx) \rangle
= \left\langle \widehat{dk}, {\textstyle \tfrac{1}{n} \sum_{p} \big(\widehat{dx}_{p}^{*} \circ \widehat{x}_{p} + \widehat{x}_{p}^{*} \circ \widehat{dx}_{p}\big)} \right\rangle \nonumber \\
& \; = {\textstyle \tfrac{1}{n} \sum_{p}} \big[ \langle \widehat{dx}_{p}, \widehat{dk}^{*} \circ \widehat{x}_{p} \rangle +
  \langle \widehat{dk} \circ \widehat{x}_{p}, \widehat{dx}_{p} \rangle \big] \nonumber \\
& \; = {\textstyle \sum_{p}} \langle \widehat{dx}_{p}, \tfrac{2}{n} \Re\{\widehat{dk}\} \circ \widehat{x}_{p} \rangle
\end{align}
to give the back-propagation map
\begin{equation}
\widehat{\nabla_{x_{p}} \ell} = \tfrac{2}{n} \widehat{x}_{p} \circ \Re\{\widehat{\nabla_{k} \ell}\} \quad \forall p \enspace .
\end{equation}
The linear map $dk, dy \mapsto d\alpha$ is identical to the single-channel case.
To find the adjoint of the map $dx, d\alpha \mapsto dw$, we examine the inner-product
\begin{align}
& \langle dw, J_{3}(dx, d\alpha) \rangle
= {\textstyle \sum_{p}} \langle \widehat{dw}_{p}, \widehat{d\alpha}^{*} \circ \widehat{x}_{p} + \widehat{\alpha}^{*} \circ \widehat{dx}_{p} \rangle \nonumber \\
& \; = \left\langle \widehat{d\alpha}, {\textstyle \sum_{p} \widehat{dw}_{p}^{*} \circ \widehat{x}_{p}} \right\rangle +
  {\textstyle \sum_{p} \langle \widehat{dw}_{p} \circ \widehat{\alpha}, \widehat{dx}_{p} \rangle} \enspace ,
\end{align}
giving the back-propagation maps
\begin{align}
\widehat{\nabla_{\alpha} \ell} & = {\textstyle \sum_{p} \widehat{x}_{p} \circ (\widehat{\nabla_{w_{p}} \ell})^{*} } \enspace , \\
\widehat{\nabla_{x_{p}} \ell} & = \widehat{\alpha} \circ \widehat{\nabla_{w_{p}} \ell} \quad \forall p \enspace .
\end{align}
Finally, combining these results gives the procedure for back-propagation in the multi-channel case
\begin{equation}
\left\{\begin{aligned}
\widehat{\nabla_{\alpha} \ell} & = {\textstyle \sum_{p} \widehat{x}_{p} \circ (\widehat{\nabla_{w_{p}} \ell})^{*} } \\
\widehat{\nabla_{y} \ell} & = \tfrac{1}{n} \widehat{k}^{-*} \circ \widehat{\nabla_{\alpha} \ell} \\
\widehat{\nabla_{k} \ell} & = -\widehat{k}^{-*} \circ \widehat{\alpha}^{*} \circ \widehat{\nabla_{\alpha} \ell} \\
\widehat{\nabla_{x_{p}} \ell} & = \widehat{\alpha} \circ \widehat{\nabla_{w_{p}} \ell} +
  \tfrac{2}{n} \widehat{x}_{p} \circ \Re\{\widehat{\nabla_{k} \ell}\} \quad \forall p \enspace .
\end{aligned}\right.
\end{equation}
Again, it is important that the computation scales only linearly with the number of channels.



\section{Hyperparameter optimization}

\begin{figure}
\centering
\includegraphics{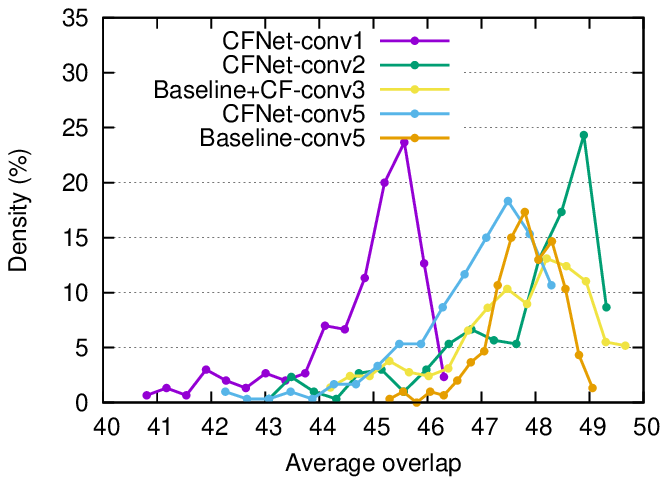}
\caption{Empirical distribution of the average overlap for the hyperparameter search.}
\label{fig:hpsearch_hist}
\end{figure}

\begin{table*}[h]
\centering
\small
\begin{tabular}{c c c c c c c c c c c c} 
\toprule
& \textbf{avg. overlap} & \textbf{best overlap} & scale step & & scale penalty & & scale l.r.\ & & win.\ weight & & template l.r.\ \\ [0.5ex]
\midrule
\textbf{CFNet-conv1}            & 44.8 & 46.5 & 1.0355 & & 0.9825 & & 0.700 & & 0.2375 & & 0.0058 \\
\textbf{CFNet-conv2}            & \bb{47.8} & \uu{49.5} & 1.0575 & & 0.9780 & & 0.520 & & 0.2625 & & 0.0050 \\
\textbf{Baseline+CF-conv3}      & \uu{47.7} & \bb{49.9} & 1.0340 & & 0.9820 & & 0.660 & & 0.2700 & & 0.0080 \\
\textbf{CFNet-conv5}            & 46.9 & 48.5 & 1.0310 & & 0.9815 & & 0.525 & & 0.2000 & & 0.0110 \\
\textbf{Baseline-conv5}         & \bb{47.8} & 49.2 & 1.0470 & & 0.9825 & & 0.680 & & 0.1750 & & 0.0102 \\
\bottomrule
\end{tabular}
\vspace{0.2cm}
\caption{Average and best overlap scores over 300 random sets of hyperparameters. Values of hyperparameters associated to the best performance are also reported. These parameters describe: the geometric step to use in scale search, the multiplicative penalty to apply for changing scale, the learning rate for updating the scale, the weight of an additive cosine window that penalizes translation, and the learning rate for the template average.}
\label{table:hps}
\end{table*}

The hyperparameters that define the simplistic tracking algorithm have a significant impact on the tracking accuracy.
These include parameters such as the penalty for changes in scale and position and the learning rate of the template average.
Choosing hyperparameters is a difficult optimization problem: we cannot use gradient descent because the function is highly discontinuous, and each function evaluation is expensive because it involves running a tracker on every sequence from multiple starting points.

For the experiments of the main paper, where we sought to make a fair comparison of different architectures, we therefore used a \emph{natural} choice of hyperparameters that were not optimized for any particular architecture.
Ideally, we would use the optimal hyperparameters for each variant, except it would have been computationally prohibitive to perform this optimization for every point in every graph in the main paper (multiple times for the points with error bars).

To achieve results that are competitive with the state-of-the-art, however, it is necessary to optimize the parameters of the tracking algorithm (on a held-out validation set).

To find optimal hyperparameters, we use random search with a uniform distribution on a reasonable range for each parameter.
Specifically, we sample 300 random vectors of hyperparameters and run the evaluation described in Section 4.1 on the 129 videos of our validation set.
Each method is then evaluated once on the test sets (OTB-2013, OTB-50 and OTB-100) using the hyperparameter vector which gave the best results on the validation set (specified in Table~\ref{table:hps}).
We emphasize that, even though the ground-truth labels are available for the videos in the benchmarks, we do not choose hyperparameters to optimize the results on the benchmarks, as this would not give a meaningful estimate of the generalization ability of the method.

Note that this random search is performed after training and is only used to choose parameters for the online tracking algorithm.
The same network is used for all random samples.
The training epoch with the best tracking results on the validation set (with natural tracking parameters) is chosen.

Figure~\ref{fig:hpsearch_hist} shows, for each method, the empirical distribution of results (in terms of average overlap) that is induced by the distribution of tracking parameters in random search.

\begin{figure*}[b]
    \centering
    \begin{subfigure}[t]{0.47\textwidth}
        \centering
        \includegraphics[width=\textwidth]{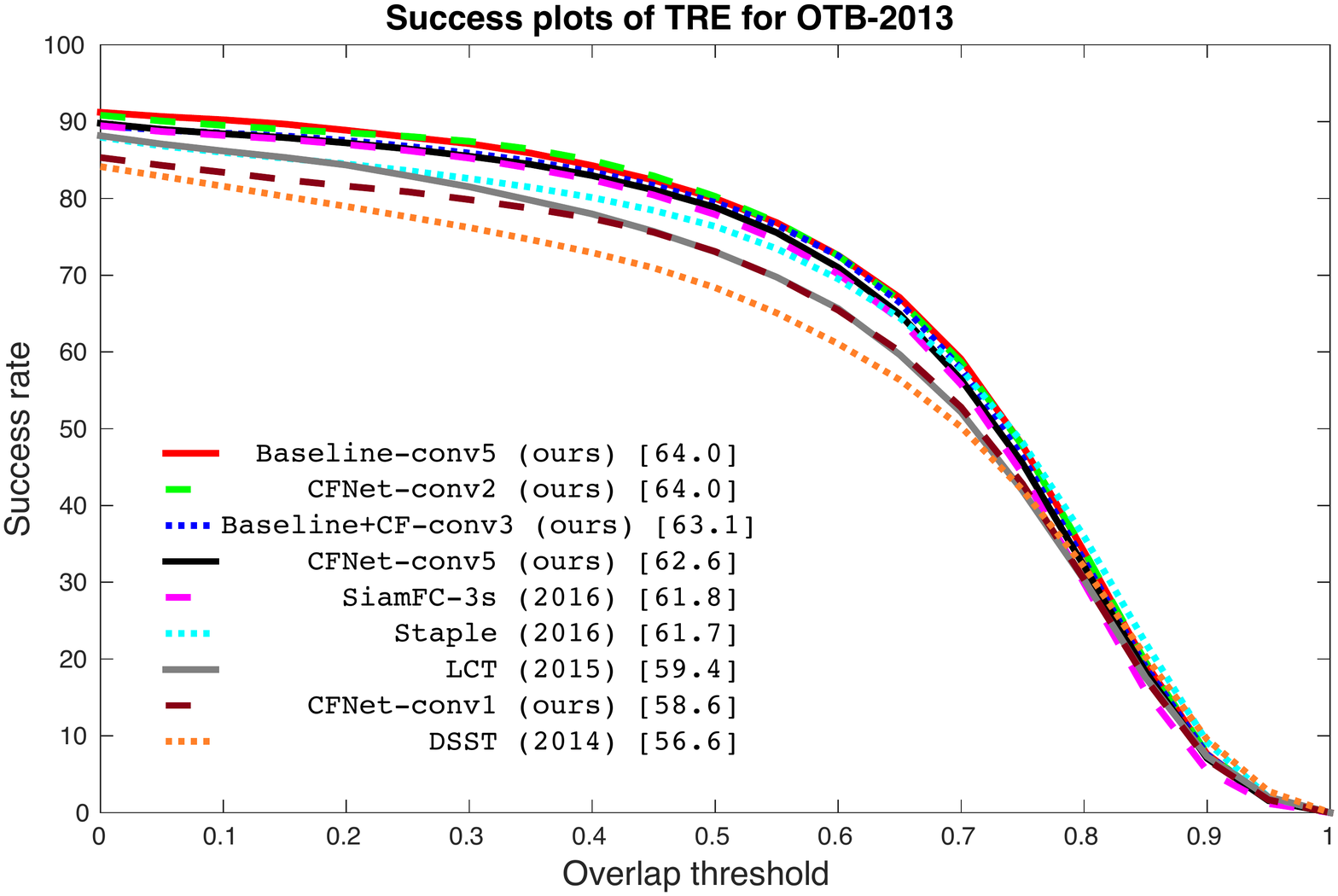}
    \end{subfigure}
    \begin{subfigure}[t]{0.47\textwidth}
        \centering
        \includegraphics[width=\linewidth]{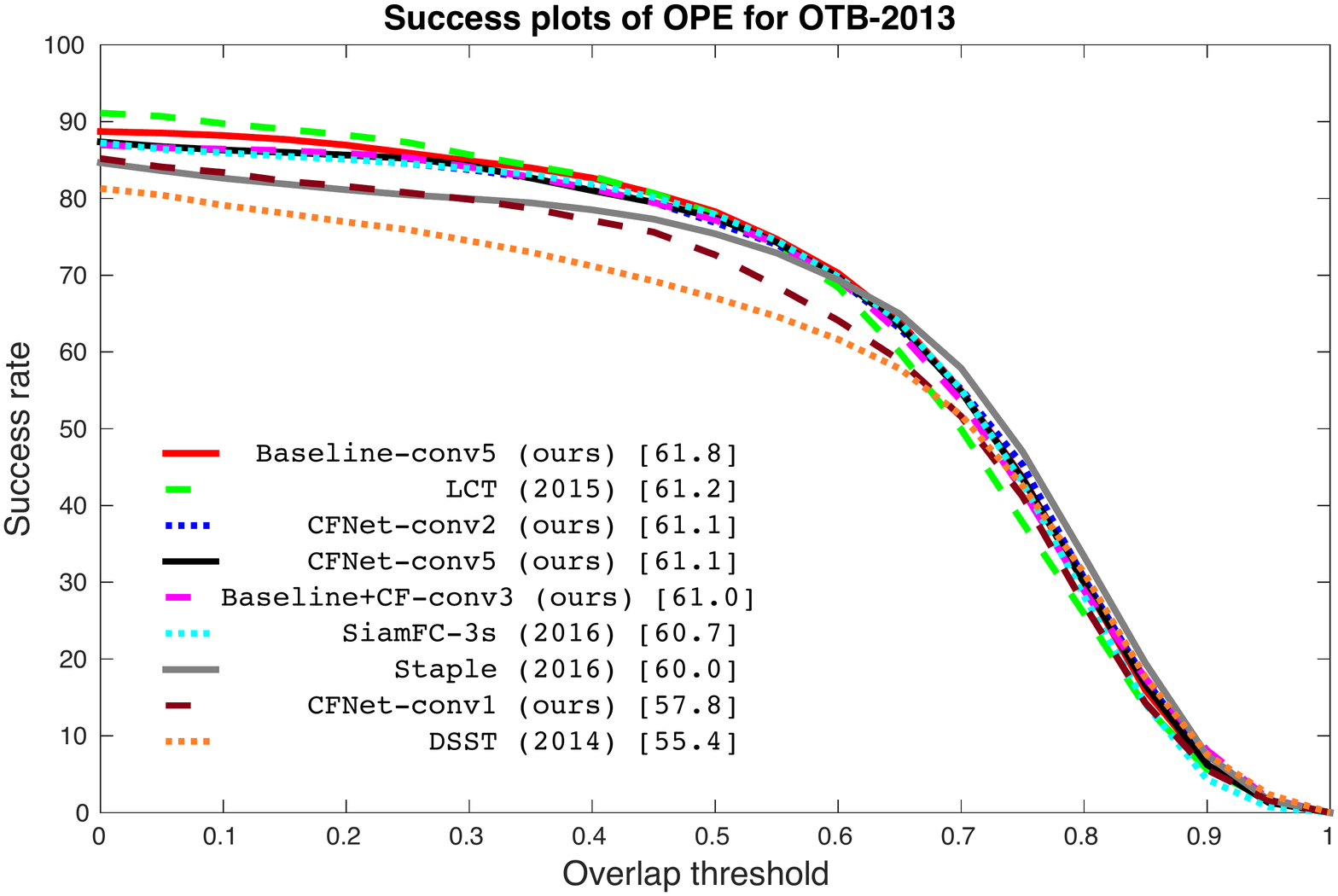}
    \end{subfigure}
    \caption{OTB-2013 success rate.}
    \label{fig:OTB-2013_overlap}
\end{figure*}

\begin{figure*}[b]
    \centering
    \begin{subfigure}[t]{0.47\textwidth}
        \centering
        \includegraphics[width=\textwidth]{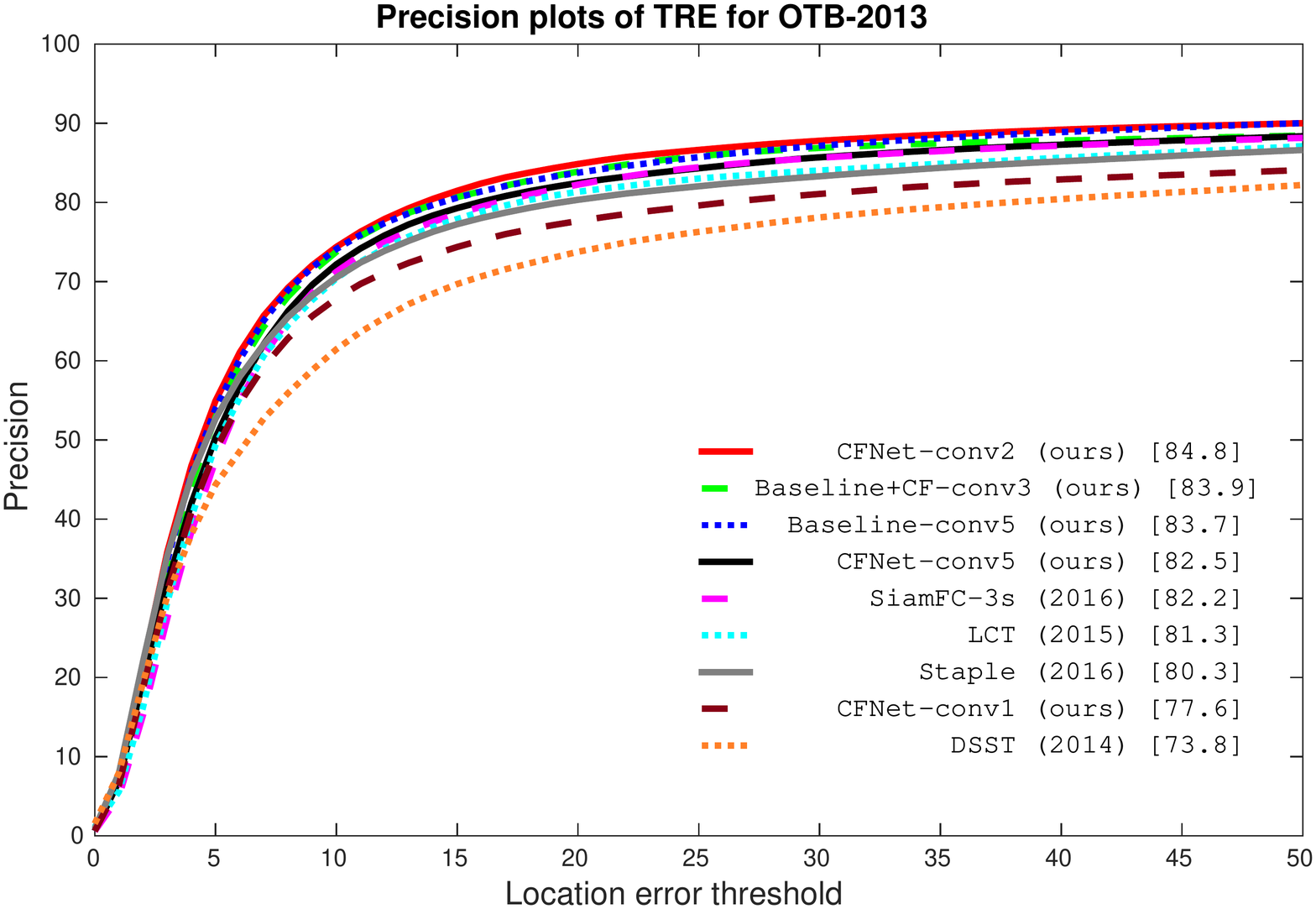}
    \end{subfigure}
    \begin{subfigure}[t]{0.47\textwidth}
        \centering
        \includegraphics[width=\linewidth]{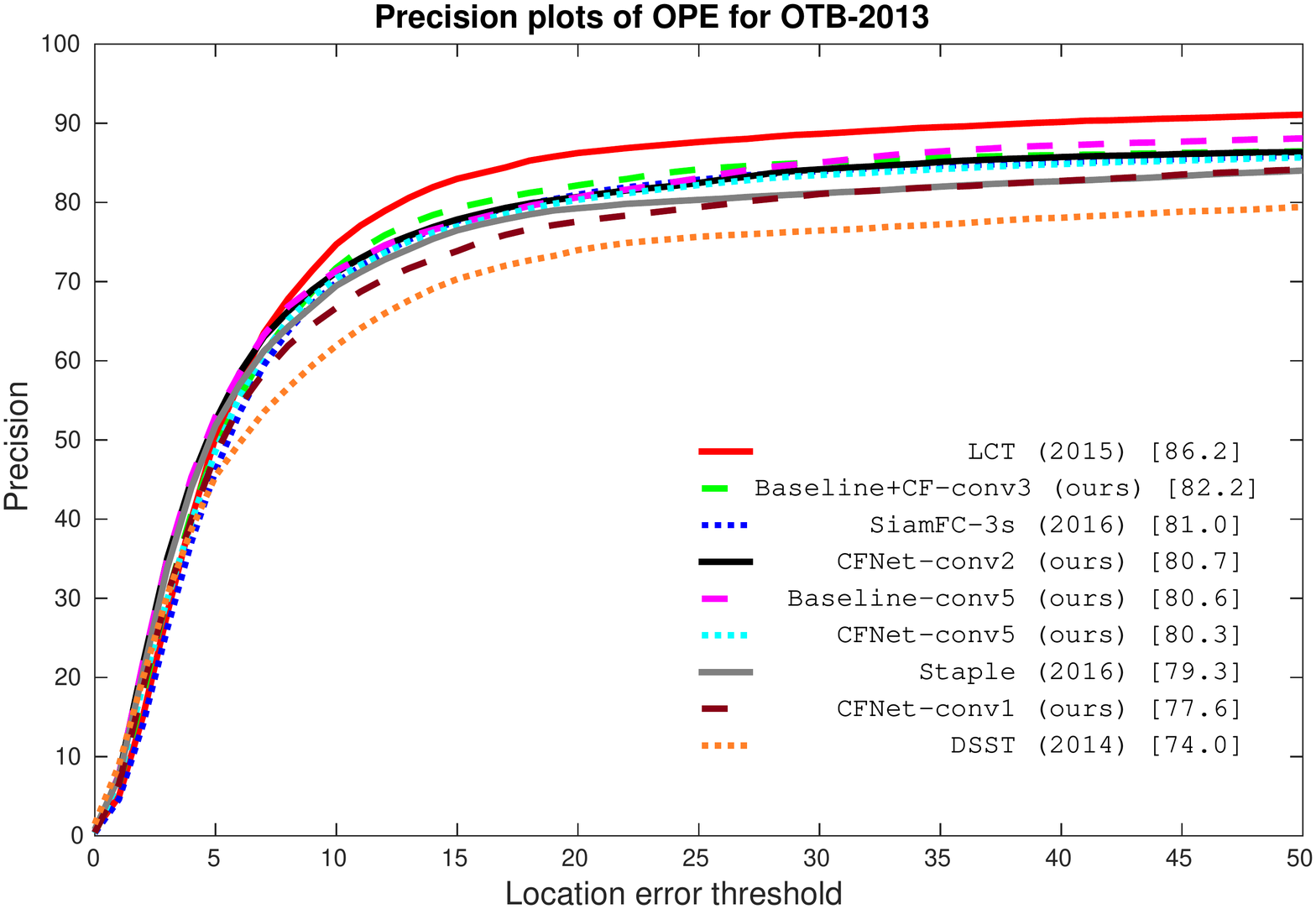}
    \end{subfigure}
    \caption{OTB-2013 precision.}
    \label{fig:OTB-2013_error}
\end{figure*}

\begin{figure*}[b]
    \centering
    \begin{subfigure}[t]{0.47\textwidth}
        \centering
        \includegraphics[width=\textwidth]{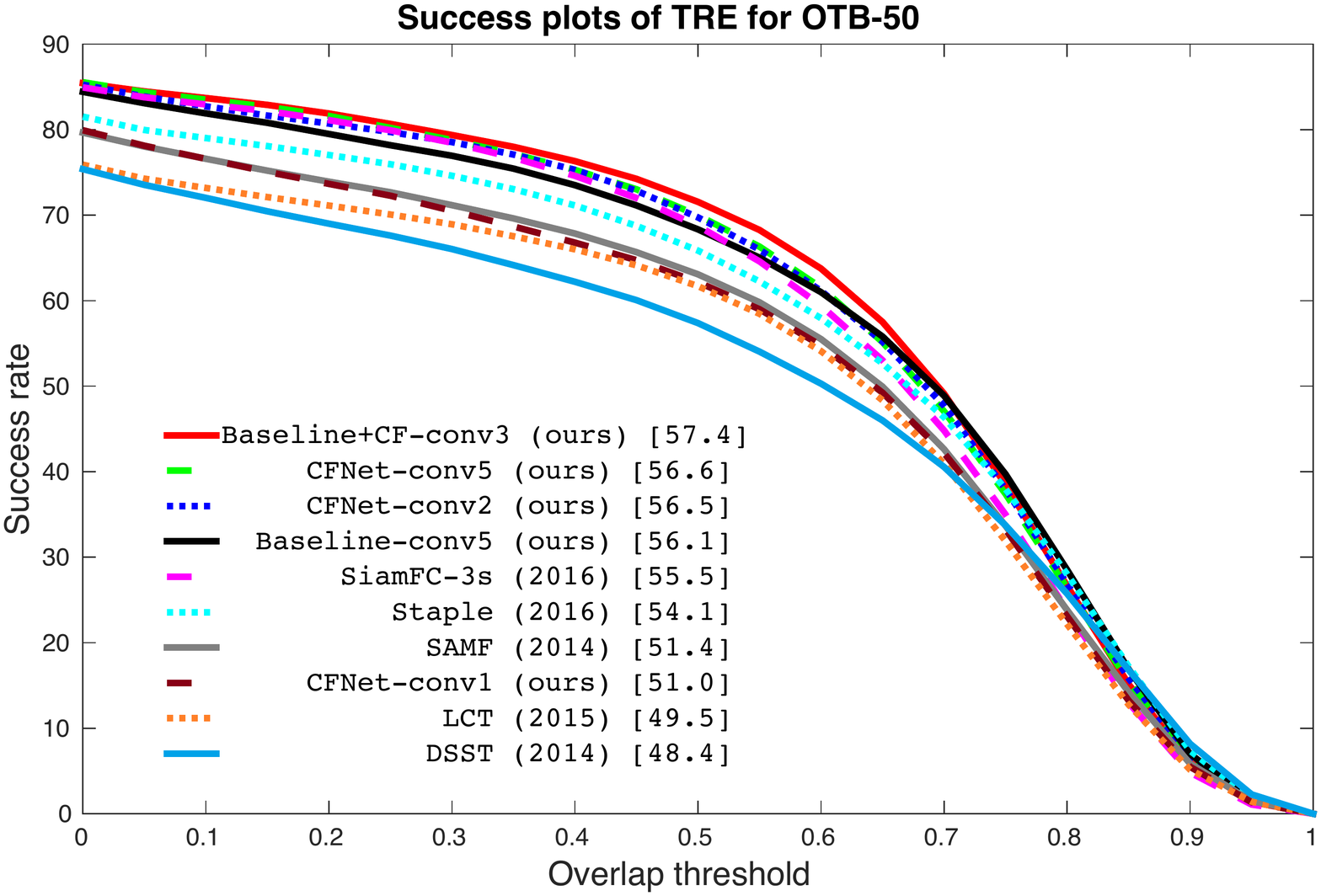}
    \end{subfigure}
    \begin{subfigure}[t]{0.47\textwidth}
        \centering
        \includegraphics[width=\linewidth]{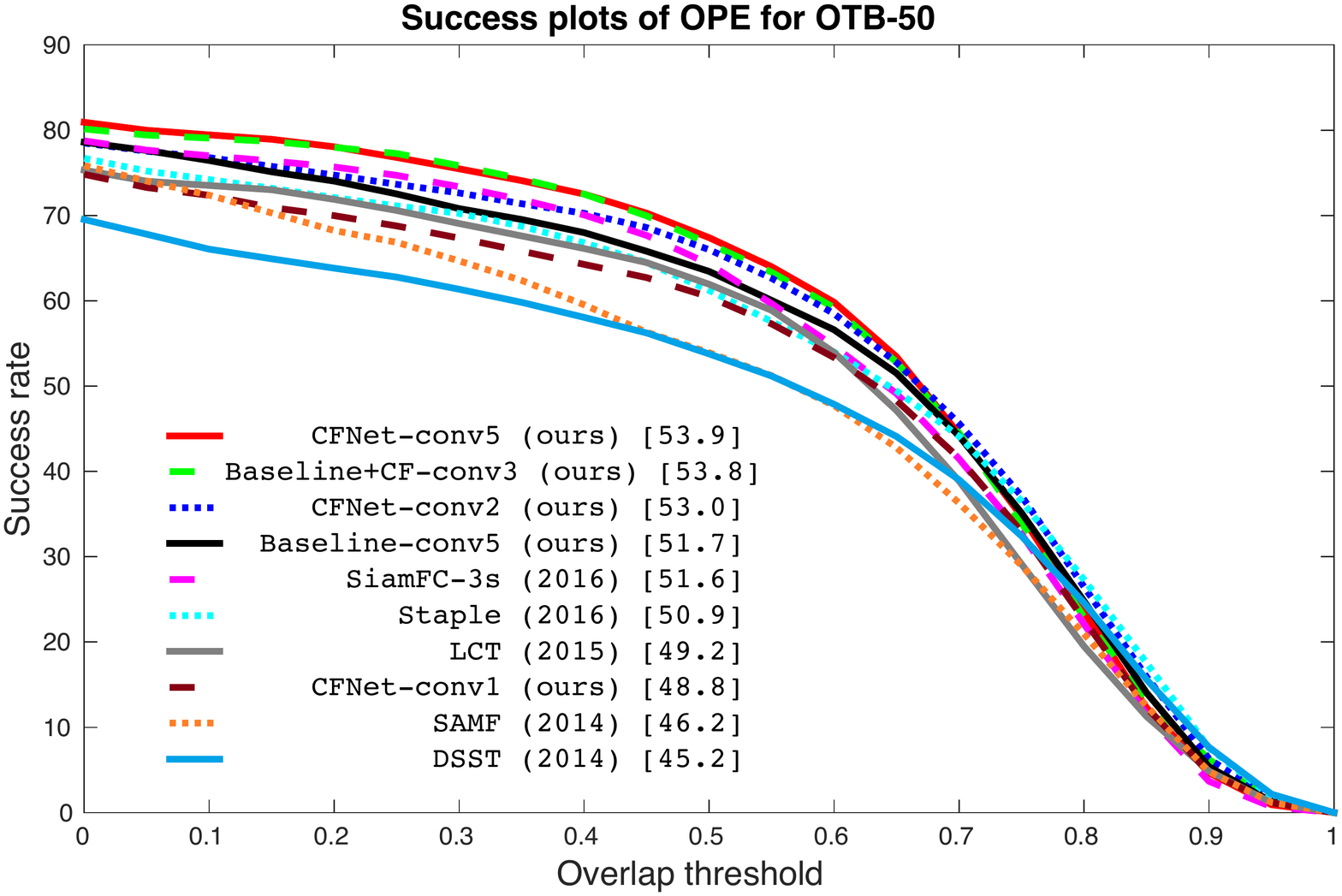}
    \end{subfigure}
    \caption{OTB-50 success rate.}
    \label{fig:OTB-50_overlap}
\end{figure*}

\begin{figure*}[b]
    \centering
    \begin{subfigure}[t]{0.47\textwidth}
        \centering
        \includegraphics[width=\textwidth]{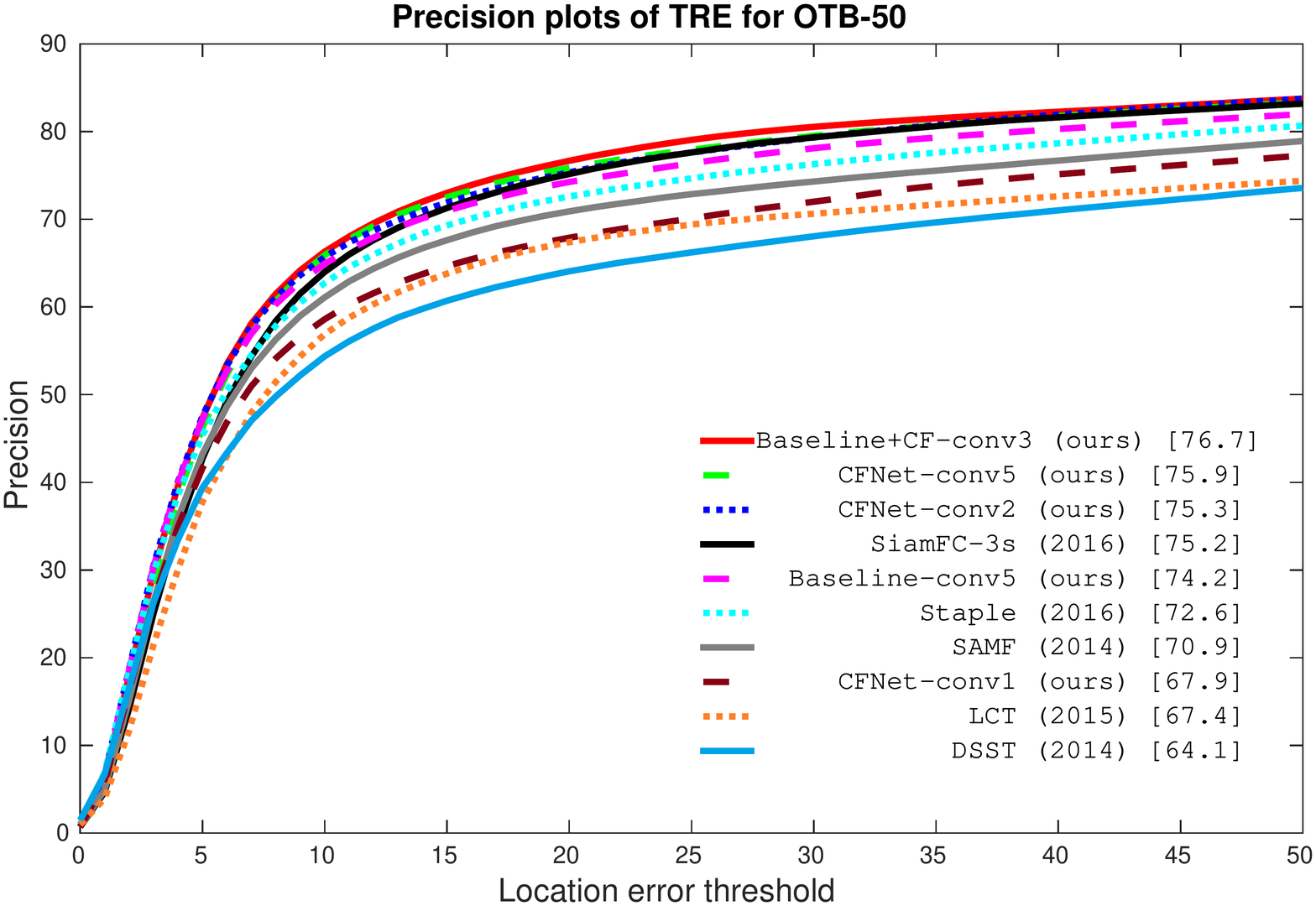}
    \end{subfigure}
    \begin{subfigure}[t]{0.47\textwidth}
        \centering
        \includegraphics[width=\linewidth]{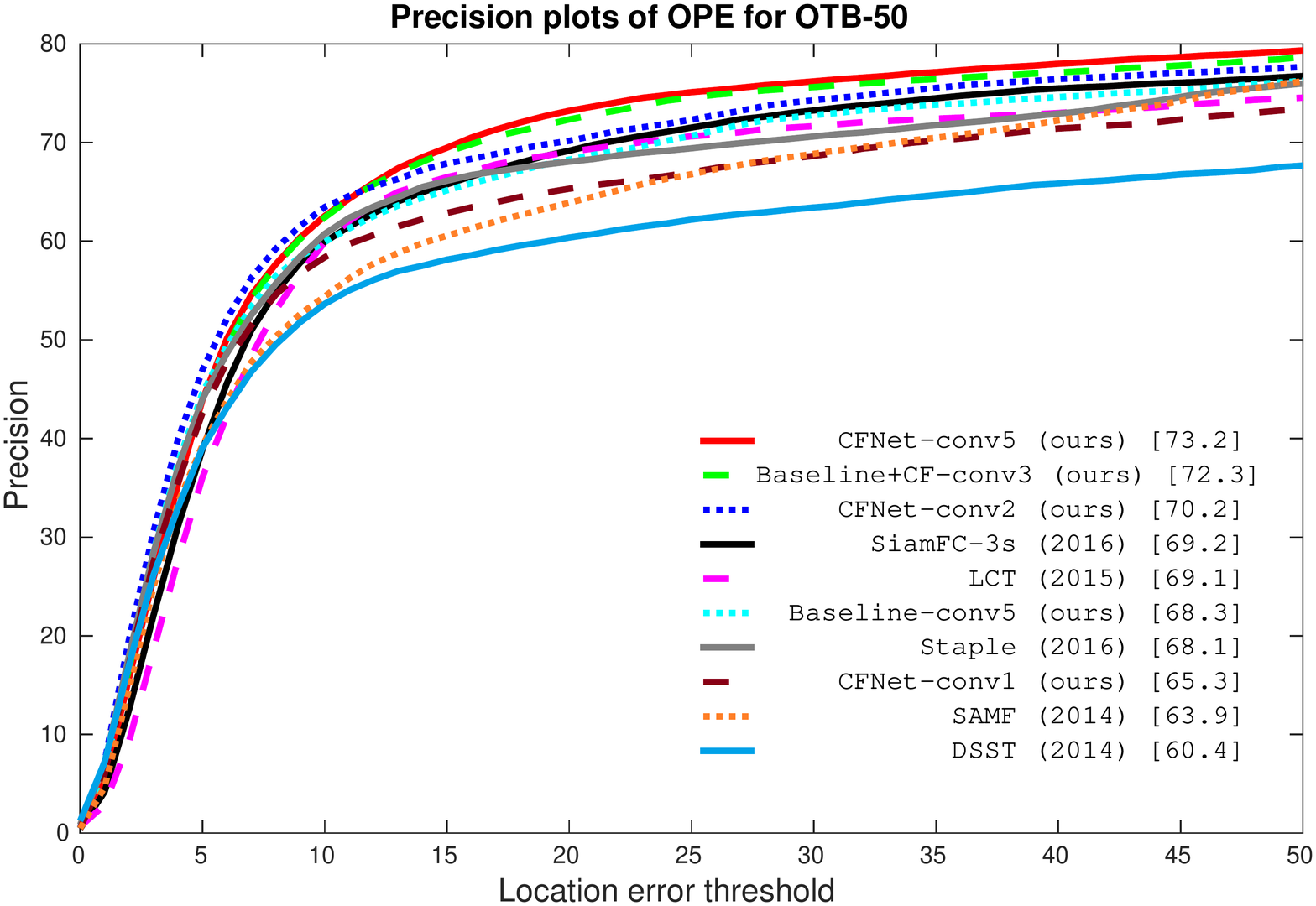}
    \end{subfigure}
    \caption{OTB-50 precision.}
    \label{fig:OTB-50_error}
\end{figure*}

\begin{figure*}[b]
    \centering
    \begin{subfigure}[t]{0.47\textwidth}
        \centering
        \includegraphics[width=\textwidth]{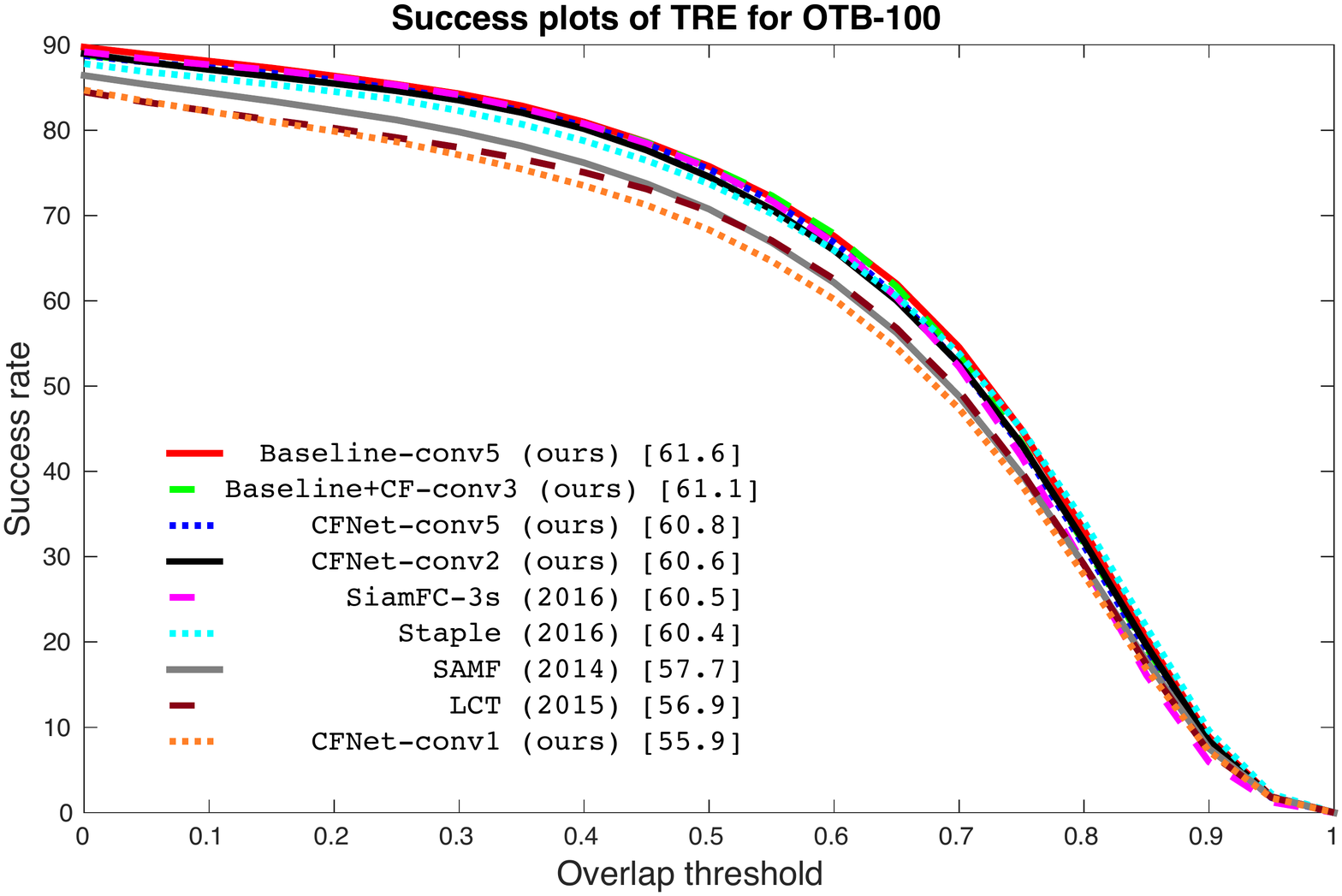}
    \end{subfigure}
    \begin{subfigure}[t]{0.47\textwidth}
        \centering
        \includegraphics[width=\linewidth]{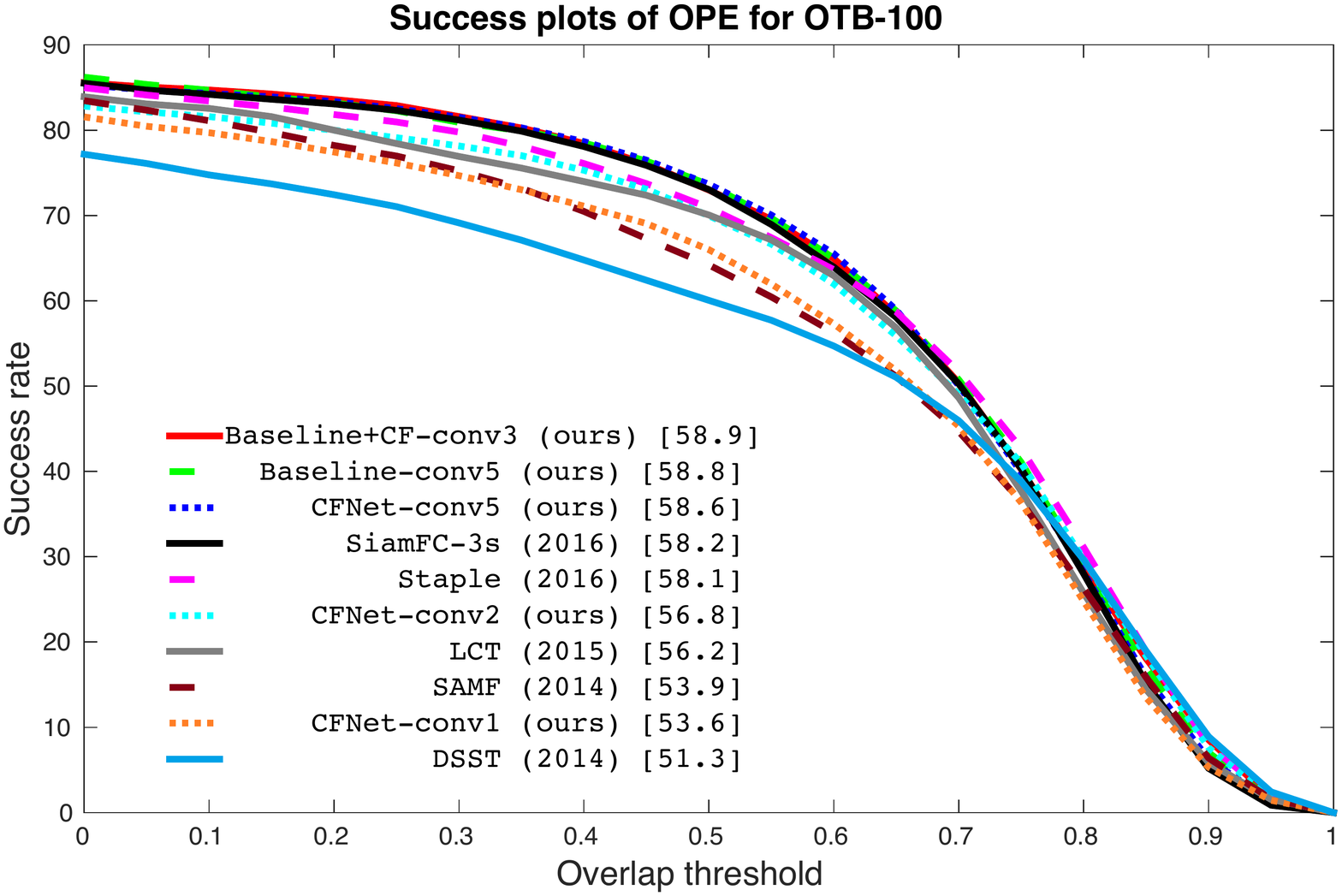}
    \end{subfigure}
    \caption{OTB-100 success rate.}
    \label{fig:OTB-100_overlap}
\end{figure*}

\begin{figure*}[b]
    \centering
    \begin{subfigure}[t]{0.47\textwidth}
        \centering
        \includegraphics[width=\textwidth]{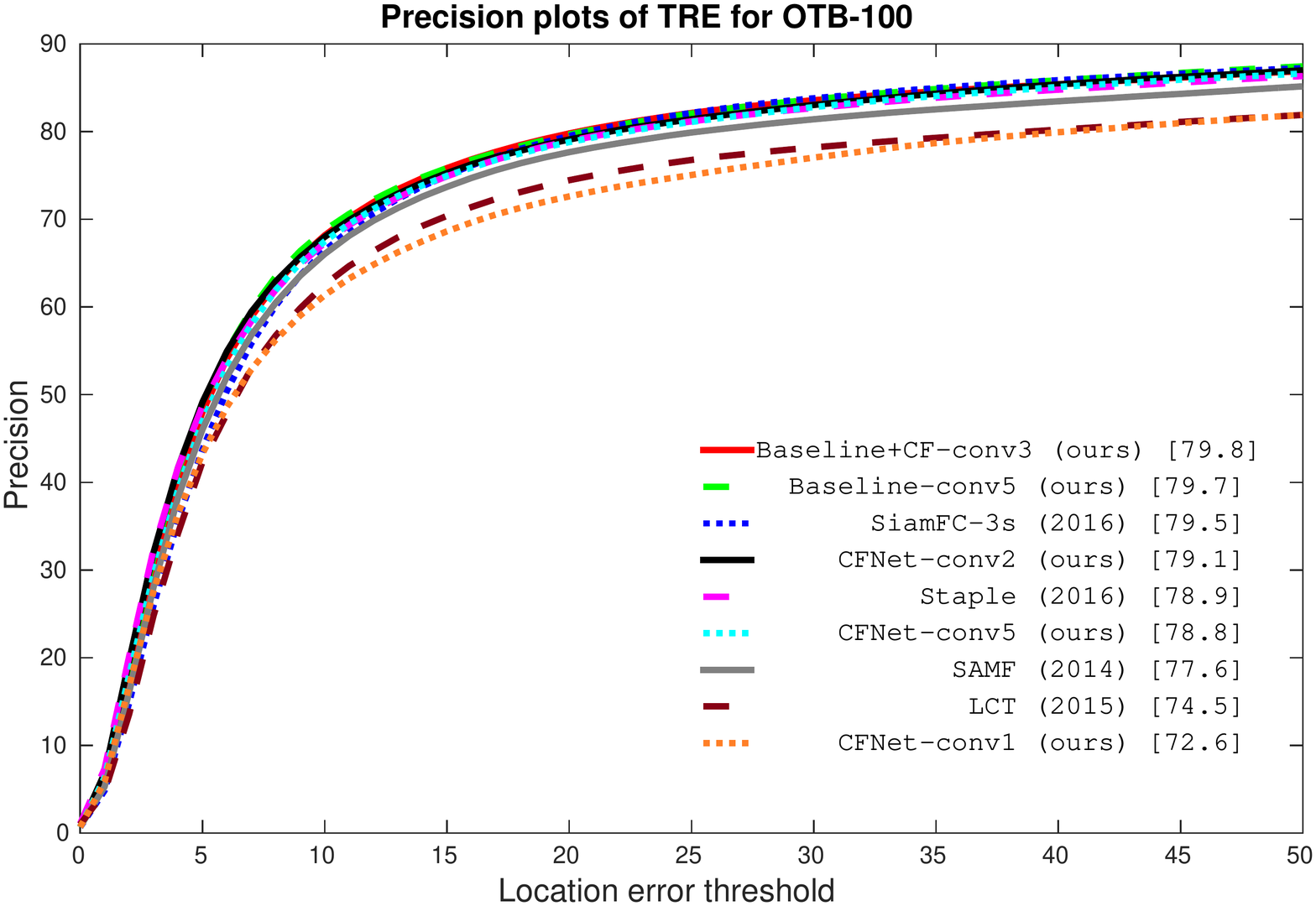}
    \end{subfigure}
    \begin{subfigure}[t]{0.47\textwidth}
        \centering
        \includegraphics[width=\linewidth]{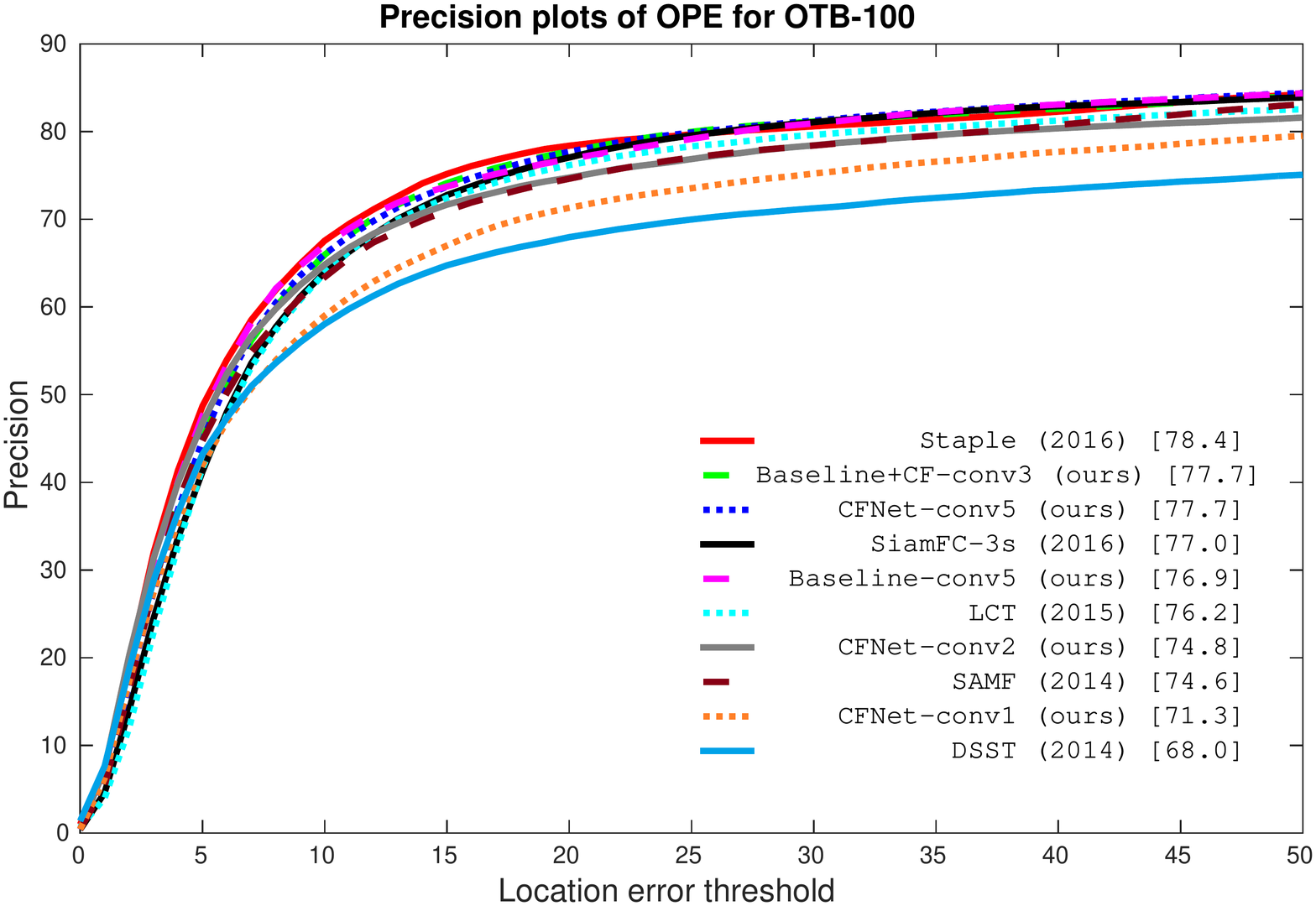}
    \end{subfigure}
    \caption{OTB-100 precision.}
    \label{fig:OTB-100_error}
\end{figure*}

\section{Detailed results on the OTB benchmarks}

Figures~\ref{fig:OTB-2013_overlap} to~\ref{fig:OTB-100_error} show the curves produced by the OTB toolkit\footnote{
  The precision plots in this version of the paper are slightly different to those in the version submitted to CVPR.
  Whereas in the CVPR version, we adopted the ``area under curve'' precision metric, here we have used the standard precision metric with a single threshold of 20 pixels.
  This has little effect on the ordering of the trackers and all observations remained valid.
} for OTB-2013/50/100, of which we presented a summary in the main paper.


\paragraph{Acknowledgements.}
This research was supported by Apical Ltd., EPSRC grant Seebibyte EP/M013774/1 and ERC grants ERC-2012-AdG 321162-HELIOS, HELIOS-DFR00200, ``Integrated and Detailed Image Understanding'' (EP/L024683/1) and ERC 677195-IDIU.

{\small
\bibliographystyle{ieee}
\bibliography{bibliography}
}


\end{document}